%% file: IEEE.tex
\documentclass[journal]{IEEEtran}
\usepackage[dvipdfmx]{graphicx}
\usepackage{cite}
\usepackage{amsmath}
\usepackage{bm}
\usepackage{here}
\usepackage{scalefnt}
\usepackage{nidanfloat}
\usepackage{mathtools}
\usepackage{amssymb}
\usepackage{mathtools}
\usepackage{color}
\usepackage{algorithm}
\usepackage{algorithmic}
\usepackage{color}
\definecolor{MYBLUE}{rgb}{0,0,1}

\begin{document}

\title{Involuntary Stabilization\\in Discrete-Event Physical Human-Robot Interaction}

\author{
	Hisayoshi~Muramatsu,~\IEEEmembership{Member,~IEEE,}
	Yoshihiro~Itaguchi,
	and~Seiichiro~Katsura,~\IEEEmembership{Member,~IEEE,}
\thanks{
	\color{blue}
	\copyright 2022 IEEE.  Personal use of this material is permitted.  Permission from IEEE must be obtained for all other uses, in any current or future media, including reprinting/republishing this material for advertising or promotional purposes, creating new collective works, for resale or redistribution to servers or lists, or reuse of any copyrighted component of this work in other works.\\\indent
	Hisayoshi Muramatsu, Yoshihiro Itaguchi, and Seiichiro Katsura, “Involuntary Stabilization in Discrete-Event Physical Human-Robot Interaction,” IEEE Transactions on Systems, Man, and Cybernetics: Systems.\\\indent
	DOI: https://doi.org/10.1109/TSMC.2022.3184960
	\color{blue}
}
\thanks{
	H. Muramatsu is with Graduate School of Advanced Science and Engineering, Hiroshima University, Higashihiroshima, Hiroshima 739-8527, Japan, (e-mail: muramatsu@hiroshima-u.ac.jp).
	Y. Itaguchi is with Faculty of Letters, Keio University, Minato-ku, Tokyo 108-8345, Japan, (e-mail: itaguchi@keio.jp).
	S. Katsura is with Department of System Design Engineering, Keio University, Yokohama, Kanagawa 223-8522, Japan, (e-mail: katsura@sd.keio.ac.jp).
}
}

\markboth{{\color{myblue} DOI: https://doi.org/10.1109/TSMC.2022.3184960} }%
{Muramatsu \color{myblue}IEEE Transactions on Systems, Man, and Cybernetics: Systems\color{myblue}}

\maketitle
\input{text/abstract}

\input{text/key}

\IEEEpeerreviewmaketitle

\input{text/section1}

\input{text/section2}

\input{text/section3}

\input{text/section4}

\input{text/section5}

\input{text/conclusion}

\input{text/thanks}
\input{text/references.bbl}

\vspace{-10mm}
\input{text/biography}

\end{document}

%% file: text/abstract.tex
\begin{abstract}
Robots are used by humans not only as tools but also to interactively assist and cooperate with humans, thereby forming physical human-robot interactions.
In these interactions, there is a risk that a feedback loop causes unstable force interaction, in which force escalation exposes a human to danger.
Previous studies have analyzed the stability of voluntary interaction but have neglected involuntary behavior in the interaction.
In contrast to the previous studies, this study considered the involuntary behavior: a human's force reproduction bias for discrete-event human-robot force interaction.
We derived an asymptotic stability condition based on a mathematical bias model and found that the bias asymptotically stabilizes a human's implicit equilibrium point far from the implicit equilibrium point and destabilizes the point near the point.
The bias model, convergence of the interaction toward the implicit equilibrium point, and divergence around the point were consistently verified via behavioral experiments under three kinds of interactions using three different body parts: a hand finger, wrist, and foot.
Our results imply that humans implicitly secure a stable and close relationship between themselves and robots with their involuntary behavior.
\end{abstract}

%% file: text/key.tex

%% file: text/section1.tex
\section{Introduction}
\IEEEPARstart{H}{uman}-robot interaction and collaboration is inevitable as robots advance \cite{Kumar2021}.
Physical interaction \cite{Li2022,Marvel2015} and context-aware interaction \cite{Liu2022,Quintas2019} between a robot and human have been studied to facilitate the robot advance.
In the physical human-robot interaction, relationship between a human and robot forms feedforward communication or feedback communication between a robot and a human.
Interaction of a robot manipulated by a human forms the feedforward communication from the human to the robot \cite{Li2018,Bowyer2015,Hassan2018}, and interaction of robotic assistance that restricts a human's movements forms the feedforward communication from the robot to the human \cite{Hassan2018,Yu2015}.
In contrast to the feedforward communications, feedback communication includes a feedback loop between a robot and a human \cite{Reed2008,Amirshirzad2019,Ficuciello2015} for interactive assistance and cooperation \cite{Patton2016,Peternel2017}.
An inherent difference between the feedforward communication and feedback communication is the feedback loop, which has a risk of destabilization, even though equilibrium points of human behavior and robot control are stable.
Thus, although the stability of human behavior and robot control can be independently evaluated in the feedforward communication, the stability cannot be independently evaluated in the feedback communication because of the feedback loop.
For the feedback communication, in addition to stability analysis of a controller, influence of the feedback loop and human behavior on the stability must be also analyzed to ensure safe interaction between a human and robot because unstable communication exposes the human to danger with force escalation.
Previous studies analyzed the stability of physical human-robot interaction, which is the feedback communication, using a model-free stability index \cite{Dimeas2016} or second-order linear time-invariant model fitting \cite{Aydin2018}.
Meanwhile, the models were not based on humans' behavioral characteristics, for example, a model of how a human's interactive force changes to cooperate with interactive force applied by a robot.
As a result, the stability analyses did not reflect the humans' behavioral characteristics.
Moreover, the studies focused on the interaction with a human's voluntary behavior and neglected the human's involuntary behavior, and the previous stability analyses neglecting the involuntary behavior were not sufficient to guarantee the stability of the human-robot interaction.
Therefore, to analyze the stability properly, it is necessary to investigate the effects of the involuntary behavior on the stability on the basis of a model of the involuntary behavior.
Although the studies on the physical human-robot interaction neglected the involuntary behavior, several studies on physical human-human interaction reported that the involuntary behavior stabilizes or destabilizes an equilibrium point of discrete-event human-human force interaction based on force reproduction.
The previous study \cite{Shergill2003} found that force escalation involuntarily occurs in the discrete-event human-human force interaction because of a human's force reproduction bias, which is caused by attenuated perception of self-generated force \cite{valles2013,Bays2006}.
On the other hand, another previous study found that force convergence occurs involuntarily in the discrete-event human-human force interaction \cite{takagi2016}, where the study interpreted the force escalation found by \cite{Shergill2003} as the convergence toward a higher equilibrium point.
Nevertheless, the bias, which is the influence of the involuntary behavior on the force reproduction, and the stability were not mathematically formulated in the previous human-human interaction studies, and they were not studied for human-robot interaction.
To analyze the stability of the human-robot interaction properly, the influence of the bias on the stability needs to be examined, and mathematical modeling of the bias and model-based stability analysis are necessary.
%

%
\input{text/Figure/Concept}
In this study, we found that the bias asymptotically stabilizes a human's implicit equilibrium point in discrete-event human-robot force interaction far from the implicit equilibrium point and destabilizes the point near the point.
The interaction is composed of alternating robot and human phases, as shown in Fig.~\ref{fig:concept}.
In the robot phase, a robot applies the same force as force applied by a human in a previous human phase, and the human perceives the force.
In the human phase, the human reproduces the force applied by the robot in the previous robot phase and applies the force to the robot, and the robot measures the applied force.
The interaction can be said to be voluntarily marginally stable because the interaction is marginally stable if there is no bias on the human's force reproduction.
Regarding the force reproduction by the human, previous studies found that the bias depends on magnitude of reproducing force \cite{walsh2011,Onneweer2013,Onneweer2016}.
We mathematically modeled the bias formulated by the implicit equilibrium point and implicit gain, as Hypothesis~1.
Using the bias model, we derived an asymptotic stability condition for the interaction, as Hypothesis~2.
Force reproduction experiments were conducted and verified Hypothesis~1, and the reproduction results and Hypothesis~2 predicted that the implicit equilibrium point of voluntarily marginally stable interaction would be asymptotically stable far from the implicit equilibrium point and unstable near the implicit equilibrium point.
Then, we experimentally evaluated the stabilities, where the interaction force converged toward his or her own implicit equilibrium point due to the local asymptotic stability and diverged around the point due to the local instability.
The bias model and stability were examined by three kinds of interactions using three different body parts: a hand finger, wrist, and foot, which indicates that the model and asymptotic stability condition may be generalizable to various discrete-event interactions using various body parts.
The contributions of this study are Hypothesis~1 for the bias model, Hypothesis~2 for the asymptotic stability condition, and the finding that the bias asymptotically stabilizes the implicit equilibrium point of the interaction far from the implicit equilibrium point and destabilizes the point near the point.

%% file: text/Figure/Concept.tex
\begin{figure}[t!]
	\begin{center}
		\includegraphics[width=\hsize]{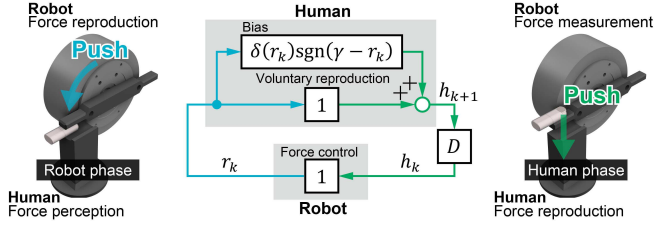}
	\caption{Discrete-event human-robot force interaction model that is voluntarily marginally stable.}\label{fig:concept}
	\end{center}
\end{figure}

%% file: text/section2.tex
\section{Modeling and Hypotheses} \label{sec:2}
\input{text/section2-1}

\input{text/section2-2}

\input{text/section2-3}

%% file: text/section2-1.tex
\subsection{Discrete-Event Human-Robot Force Interaction} \label{sec:2-1}
Consider the discrete-event human-robot force interaction:
\begin{align}
	\label{eq:DEHRFI}
	\left\{
	\begin{array}{l}
		h_{k+1}=[1+U(r_k)]r_k\\
		r_k=h_k
	\end{array}
	\right.,\ k=0,\ 1,\ 2,\ 3,\ \ldots\ ,
\end{align}
which is depicted in Fig.~\ref{fig:concept}.
A human and robot apply force to each other.
The variable $h_{k+1}$ denotes the interaction force voluntarily reproduced by the human from the interaction force $r_k$.
For the reproducing human, the interaction forces $h_{k+1}$ and $r_k$ are subjectively the same, but there is the bias $U(r_k)$ involuntarily.
The variable $r_k$ denotes the interaction force reproduced by the robot, which is assumed to be precise reproduction.
The shift operator $D$ is defined to satisfy $h_{k+1}=Dh_k$.
The signs of the interaction forces $h_k$ and $r_k$, which are the directions of the force applications, are assumed to be positive
\begin{align}
	\label{eq:DEHRFIcondition}
	h_k,\ r_k>0.
\end{align}
The interaction is voluntarily marginally stable, which indicates that the interaction is marginally stable $h_{k+1}=r_k=h_k$ if there is no bias $U(r_k)=0$.
The robot and human phases are alternately performed in the interaction.

%% file: text/section2-2.tex
\subsection{Hypotheses} \label{sec:2-2}
We set up two hypotheses.
One is for a mathematical model of the bias.\\

\noindent
\textbf{Hypothesis 1.} The bias $U(r_k)$ of the force reproduction by a human obeys
\begin{align}
	\label{eq:Bias}
	U(r_k)=\delta(r_k)\mathrm{sgn}(\gamma-r_k),
\end{align}
where $\delta(r_k)$ and $\gamma$ are the implicit gain and implicit equilibrium point, respectively.
\\

\noindent
The model reflects the results of the previous studies that the bias depends on the reproducing force level \cite{walsh2011,Onneweer2013,Onneweer2016,takagi2016}.
To model the bias, we defined and used the implicit gain $\delta(r_k)$ and the implicit equilibrium point $\gamma$, which are related to the amplification and the sign of the bias, respectively, and characterize the convergence of the interaction.
The other hypothesis is an asymptotic stability condition for the interaction.\\

\noindent
\textbf{Hypothesis 2.} The implicit equilibrium point $\gamma$ of the discrete-event human-robot force interaction is asymptotically stable if the asymptotic stability condition:
\begin{align}
	\label{eq:AScondition}
	[r_k\delta(r_k)-2|\gamma-r_k|]\delta(r_k)<0\ \mathrm{if}\ r_k\neq\gamma
\end{align}
is satisfied.
\\

\noindent
The asymptotic stability condition \eqref{eq:AScondition} is derived from the models \eqref{eq:DEHRFI} and \eqref{eq:Bias}.
For asymptotic stability analysis, we defined the evaluation value $E$ with the left-hand side of \eqref{eq:AScondition} as
\begin{align}
	\label{eq:Evalue}
	E \coloneqq [r_k\delta(r_k)-2|\gamma-r_k|]\delta(r_k).
\end{align}
The symbol $\coloneqq$ stands for the definition.

%% file: text/section2-3.tex
\subsection{Asymptotic Stability Condition} \label{sec:2-3}
An error $e_k$ is defined by
\begin{align}
	\label{eq:error:DEF}
	e_k \coloneqq \gamma-r_k,
\end{align}
and the asymptotic stability of the origin of the error $e_k$ is equivalent to that of the implicit equilibrium point $\gamma$.
The origin of $e_k$ is asymptotically stable if the discrete-time Lyapunov stability condition
\begin{subequations}
	\label{eq:ASC:1}
\begin{align}
	\label{eq:ASC:1a}
	\Delta V [e_k] &\coloneqq e_{k+1}^2 - e_{k}^2=0\ \mathrm{if}\ e_k=0\\
	\label{eq:ASC:1b}
	\Delta V [e_k] &\coloneqq e_{k+1}^2 - e_{k}^2<0\ \mathrm{if}\ e_k\neq0
\end{align}
\end{subequations}
is satisfied.
According to \eqref{eq:DEHRFI} and \eqref{eq:error:DEF}, the error can be expressed as
\begin{align}
	\label{eq:ASC:2}
	e_k=\gamma-r_k=\gamma-h_k.
\end{align}
If the origin of $e_k$ is asymptotically stable, the square error $e_{k+1}^2$ decreases from the previous square error $e_k^2$ toward zero, and the error remains at zero if the error is zero.
This indicates that the interaction forces $r_k$ and $h_k$ converge at the implicit equilibrium point $\gamma$.
The discrete-time Lyapunov stability condition \eqref{eq:ASC:1} holds if the asymptotic stability condition \eqref{eq:AScondition} is satisfied, which is proved below.
The function $\Delta V[e_k]$ is transformed using the interaction model \eqref{eq:DEHRFI} into
\begin{align}
	\label{eq:ASC:3}
	\Delta V[e_k]=(\gamma-h_{k+1})^2-e_k^2=[\gamma-(1+U)r_k]^2-e_k^2.
\end{align}
Using \eqref{eq:Bias} and \eqref{eq:ASC:2}, \eqref{eq:ASC:3} is further transformed as follows
\begin{align}
	\label{eq:ASC:4}
	\Delta V[e_k]&=[\gamma-(1+U)(\gamma-e_k)]^2-e_k^2\notag\\
	&=[e_k-U(\gamma-e_k)]^2-e_k^2\notag\\
	&=-2Ue_k(\gamma-e_k)+U^2(\gamma-e_k)^2,
\end{align}
where $Ue_k$ and $U^2$ can be expressed using \eqref{eq:Bias} as
\begin{align}
	\label{eq:ASC:5}
	Ue_k=\delta(r_k)|\gamma-r_k|,\
	U^2=\delta^2(r_k)\mathrm{sgn}^2(e_k).
\end{align}
From \eqref{eq:ASC:5}, \eqref{eq:ASC:4} becomes
\begin{align}
	\label{eq:ASC:6}
	&\Delta V[e_k]=\notag\\
	&-2\delta(r_k)|\gamma-r_k|(\gamma-e_k)+\delta^2(r_k)\mathrm{sgn}^2(e_k)(\gamma-e_k)^2.
\end{align}
Finally, $\Delta V[e_k]$ is expressed by eliminating the error $e_k$ of \eqref{eq:ASC:6} as
\begin{align}
	\label{eq:ASC:7}
	&\Delta V[\gamma-r_k]=\notag\\
	&-2\delta(r_k)|\gamma-r_k|r_k + \delta^2(r_k)\mathrm{sgn}^2(\gamma-r_k)r_k^2.
\end{align}
The condition \eqref{eq:ASC:1a} is satisfied because
\begin{align}
	\label{eq:ASC:8}
	\Delta V[0]=0,
\end{align}
where $\mathrm{sgn}(0)=0$.
Moreover, using $0<r_k$ and $\mathrm{sgn}^2(\gamma-r_k)=1$ if $e_k\neq0$, the condition \eqref{eq:ASC:1b} is satisfied if the condition \eqref{eq:AScondition} is satisfied.
Therefore, the discrete-time Lyapunov stability condition \eqref{eq:ASC:1} holds if the asymptotic stability condition \eqref{eq:AScondition} is satisfied.
Furthermore, the implicit equilibrium point $\gamma$, which equals the origin of $e_k$, is asymptotically stable if the asymptotic stability condition \eqref{eq:AScondition} is satisfied.

%% file: text/section3.tex
\section{Method} \label{sec:3}
\input{text/Figure/ExpSys}
\input{text/section3-1}

\input{text/section3-2}

\input{text/section3-3}

%% file: text/Figure/ExpSys.tex
\begin{figure}[t!]
	\begin{center}
		\includegraphics[width=\hsize]{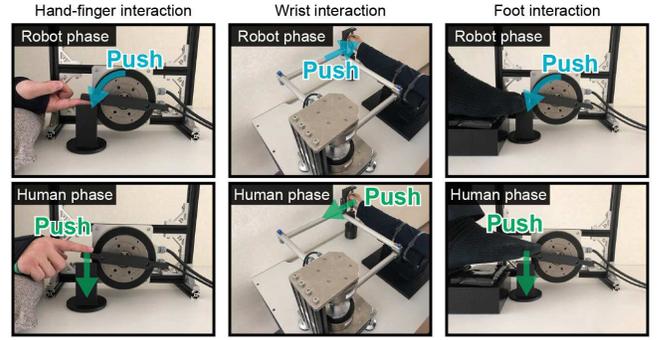}
	\caption{Hand finger, wrist, and foot interactions using a right hand index finger, right wrist, and right foot big toe, respectively.}\label{fig:ExpSys}
	\end{center}
\end{figure}

%% file: text/section3-1.tex
\subsection{Objectives and Study Design} \label{sec:3-1}
We aimed to demonstrate Hypotheses 1 and 2.
To this end, we designed four objectives: verification of the bias model (Experiment 1-A), stability prediction based on force reproduction results (Experiment 1-A), validation of the predicted stability with human-robot interaction results (Experiment 1-B), and exploration of the generalizability (Experiments 2-A, 2-B, 3-A, and 3-B).
The first objective was for Hypothesis 1, the second and third objectives were for Hypothesis 2, and the last objective was for both the hypotheses.

%% file: text/section3-2.tex
\subsection{Experimental Design} \label{sec:3-2}
\subsubsection{Robot System, Measurements, and Control}
The hand-finger and foot interactions used the same one-degree-of-freedom robot (Fig.~\ref{fig:ExpSys} left and right), and the wrist interaction used the two-degrees-of-freedom parallel robot (Fig.~\ref{fig:ExpSys} middle).
Both the robots used direct drive servomotors: SGMCS-02BDC41 from YASKAWA ELECTRIC CORPORATION.
Motor angle was measured by rotary encoders, and torque was calculated using a reaction force observer \cite{1993_Murakami_FC, katsura2007modeling}.
Each experiment consisted of the robot phase and human phase.
In the robot phase, force control was implemented in the robot systems with a disturbance observer \cite{2019_Sariyildiz_DOB, 2015_Sariyildiz_DOB}, where the disturbance observer was used to realize precise control by compensating for disturbances.
The force controller was
\begin{align}
	\label{eq:}
	I_{\mathrm{a}}(s)=\frac{J_{\mathrm{n}}}{K_{\mathrm{tn}}}  C_{\mathrm{f}} (T^{\mathrm{cmd}}(s)-T^{\mathrm{res}}(s) )+I^{\mathrm{cmp}}(s),
\end{align}
which is analyzed in \cite{2015_Sariyildiz_FC}, including its stability.
The variables $I_\mathrm{a}$, $J_\mathrm{n}$, $K_\mathrm{tn}$, $C_\mathrm{f}$, $T^\mathrm{cmd}$, $T^\mathrm{res}$, $I^{\mathrm{cmp}}$, and $s$ denote the current input for the motors, nominal inertia, nominal torque constant, force gain, torque command, torque response, compensation current, and Laplace operator, respectively.
The compensation current $I^{\mathrm{cmp}}$ was provided by the disturbance observer as
\begin{align}
	\label{eq:}
	I^{\mathrm{cmp}}(s)=\frac{g}{s+g}\left[K_{\mathrm{tn}}I_{\mathrm{a}}(s)-J_{\mathrm{n}}\left(\frac{g_{\mathrm{pd}}}{s+g_{\mathrm{pd}}}s\right)^2\theta^{\mathrm{res}}(s)\right],
\end{align}
where $g$, $g_\mathrm{pd}$, and $\theta^\mathrm{res}$ denote the cutoff frequency for the disturbance observer, cutoff frequency for the pseudo differentiation, and angle response, respectively.
In the human phase, zero-command position control was implemented in the robot systems.
The position controller was
\begin{align}
	\label{eq:}
	I_{\mathrm{a}}(s)=\frac{J_{\mathrm{n}}}{K_{\mathrm{tn}}}\left[K_{\mathrm{p}}+K_{\mathrm{v}}\frac{g_{\mathrm{pd}}}{s+g_{\mathrm{pd}}}s\right](0-\theta^{\mathrm{res}}(s))+I^\mathrm{cmp}(s),
\end{align}
where $K_\mathrm{p}$ and $K_\mathrm{v}$ are the proportional and derivative gains, respectively.
The workspace position and force commands were calculated into the joint space angle and torque commands, and the controls were implemented in the joint space.
The controllers were implemented on a PC using the real-time application interface (RTAI) for Linux and the advanced robot control system (ARCS) with sampling time at 0.1 ms.
\subsubsection{Experiment 1}
Twelve right-handed participants from 20 to 26 years old took part in Experiments 1-A (force reproduction) and 1-B (interaction).
Experiment 1 employed hand-finger interaction (Fig.~\ref{fig:ExpSys} left).
The participants wore a wrist supporter to fix their wrist movements.
The motor applied force to a right hand index finger for 2 s in the robot phase.
The participants were instructed to perceive the force without pushing back the motor in the robot phase, while the participants did not know how the robot applies force to the participants so that the participants were not aware of their involuntary behavior.
In the human phase, the participants were instructed to apply the same force using the index finger for 2 s.
The steady-state force, which was the force during the last 1 s of each phase, was measured as the interaction forces $h_k$ and $r_k$.

Experiment 1-A: The force reproduction experiment had fifty sets of two robot and human phases.
In the robot phase, the motor applied one of ten forces ($1,\ 2,\ \ldots,\ 10\ \mathrm{N}$).
Each force was selected five times, and the order of the applied forces was randomly determined.
Then, a participant reproduced the force in the human phase.

Experiment 1-B: The interaction experiment had ten sets of forty robots and human phases.
Each interaction was initiated by a robot phase, and the ten multiple interactions started with ten different initial forces randomly selected from $1,\ 2,\ \ldots,\ 10\ \mathrm{N}$ without repetition.
In the other subsequent robot phases, the motor applied force that was the same as the force applied by a participant in a previous human phase.
The robot and human phases were alternately conducted.

We designed the experiments in accordance with the previous study \cite{Shergill2003}, but our interaction experiments used forty robot and human phases, which was greater than those of \cite{Shergill2003} because the interactions of the previous study might not reach steady-state and showed only transient interactions in \cite{Shergill2003}.
The number of the phases was determined by pilot experiments to observe steady-state interactions and reduce participants' fatigue.
\subsubsection{Experiment 2}
Six right-handed participants from 20 to 25 years old took part in Experiments 2-A (force reproduction) and 2-B (interaction), where four of the participants took part in Experiment 1 also.
Experiment 2 employed wrist interaction (Fig.~\ref{fig:ExpSys} middle).
The right arm of each participant was fixed on an armrest, and the participant gripped the end-effector of the robot.
The robot applied force through the gripped end-effector for 3 s in the robot phase.
The participants were instructed to perceive the force without pushing back the end-effector in the robot phase.
They were instructed to apply the same force using their wrist for 3 s in the human phase.
Similar to Experiments 1-A and 1-B, Experiments 2-A and 2-B were force reproduction and interaction experiments, respectively.
The applied forces in Experiment 2-A and the ten initial forces in Experiment 2-B were randomly selected from $0.5,\ 1,\ \ldots,\ 5\ \mathrm{N}$ for the robot phases.
The steady-state force, which was the force during the last 1 s of a phase, was measured as the interaction forces $h_k$ and $r_k$.
\subsubsection{Experiment 3}
The same six right-handed participants from Experiment 2 took part in Experiments 3-A (force reproduction) and 3-B (interaction).
Experiment 3 employed foot interaction (Fig.~\ref{fig:ExpSys} right).
The motor applied force to a right foot big toe for 2 s in the robot phase.
The participants were instructed to perceive the force without pushing back the motor in the robot phase and applied the same force using their big toe for 2 s in the next human phase.
Similar to Experiments 1-A, 1-B, 2-A, and 2-B, Experiments 3-A and 3-B were force reproduction and interaction experiments, respectively.
The applied forces in Experiment 3-A and the ten initial forces in Experiment 3-B were randomly selected from $2.5,\ 5,\ \ldots,\ 25\ \mathrm{N}$ for the robot phases.
The steady-state force, which was the force in the last 1 s of a phase, was measured as the interaction forces $h_k$ and $r_k$.
\subsubsection{Outlier}
In Experiment 1-B, one participant whose mean of the final absolute errors $\epsilon_\mathrm{h20}$ was greater than the other participants' mean+10$\times$SD was judged as an outlier.
Accordingly, the participant's data were eliminated from the results.
Additionally, Participant 2 (P2) of Experiment 2-B was not used to evaluate the normalized absolute errors because the interaction force might not converge in forty trials owing to the participant's large implicit equilibrium point ($16.42\ \mathrm{N}$).

%% file: text/section3-3.tex
\subsection{Statistical Hypothesis Testing} \label{sec:3-3}
This study used statistical hypothesis testing to evaluate the stability, the convergence of the normalized absolute errors, and comparison of the fitting errors and implicit equilibrium points.
The stability evaluation in Sections~\ref{sec:4-2} and \ref{sec:4-4} used a one-sided Student's t-test, which is a one-sample test for the evaluation values $E$ of a participant.
The null hypothesis was $\mu\geq0$, and the alternative hypothesis was $\mu<0$, where $\mu$ is the mean of the evaluation values $E$ for the asymptotic stability criteria.
The significance level was set at 0.05.
The convergence evaluation in Sections~\ref{sec:4-3} and \ref{sec:4-4} used a one-sided paired t-test by classifying each participant's ten interactions into Groups i--x, and we tested the paired initial and final errors of each group.
A group whose initial normalized absolute error $\epsilon_{\mathrm{r}0}$ was not in the unstable region was tested with the null hypothesis: $\bar{\epsilon}_\mathrm{r0}\leq\bar{\epsilon}_\mathrm{h20}$ and the alternative hypothesis: $\bar{\epsilon}_\mathrm{r0}>\bar{\epsilon}_\mathrm{h20}$.
In contrast, a group whose initial normalized absolute error $\epsilon_{\mathrm{r}0}$ was in the unstable region was tested with the null hypothesis: $\bar{\epsilon}_\mathrm{r0}\geq\bar{\epsilon}_\mathrm{h20}$ and the alternative hypothesis: $\bar{\epsilon}_\mathrm{r0}<\bar{\epsilon}_\mathrm{h20}$.
$\bar{\epsilon}_\mathrm{r0}$ and $\bar{\epsilon}_\mathrm{h20}$ are the means of the normalized absolute errors $\epsilon_\mathrm{r0}$ and $\epsilon_\mathrm{h20}$, respectively.
The significance level was set at 0.05.
The discussion of Hypothesis 1 in Section~\ref{sec:5-1} used a one-sided Welch's t-test, which was used to compare the fitting errors and implicit equilibrium points between the three tasks with different variances.
For the fitting errors of the hand-finger $\mu_\mathrm{H}$, wrist $\mu_\mathrm{W}$, and foot $\mu_\mathrm{F}$ interactions, the null hypotheses were $\mu_\mathrm{F}\leq\mu_\mathrm{H}$ and $\mu_\mathrm{W}\leq\mu_\mathrm{H}$, and the alternative hypotheses were $\mu_\mathrm{F}>\mu_\mathrm{H}$ and $\mu_\mathrm{W}>\mu_\mathrm{H}$, respectively.
For the implicit equilibrium points of the hand-finger $\mu_\mathrm{H}$, wrist $\mu_\mathrm{W}$, and foot $\mu_\mathrm{F}$ interactions, the null hypotheses were $\mu_\mathrm{F}\leq\mu_\mathrm{H}$ and $\mu_\mathrm{W}\leq\mu_\mathrm{H}$, and the alternative hypotheses were $\mu_\mathrm{F}>\mu_\mathrm{H}$ and $\mu_\mathrm{W}>\mu_\mathrm{H}$, respectively.
The significance level was set at 0.05.

%% file: text/section4.tex
\section{Results} \label{sec:4}
\input{text/Figure/Exp1-A}

\input{text/section4-1}

\input{text/section4-2}
\input{text/Figure/Exp1-B}

\input{text/section4-3}

\input{text/Figure/Exp2}

\input{text/Figure/Exp3}

\input{text/section4-4}

\input{text/section4-5}

\input{text/Figure/Comparison}

%% file: text/Figure/EXP1-A.tex
\begin{figure*}[t]
	\begin{minipage}{0.83\hsize}
		\begin{center}
			\includegraphics[height=55mm]{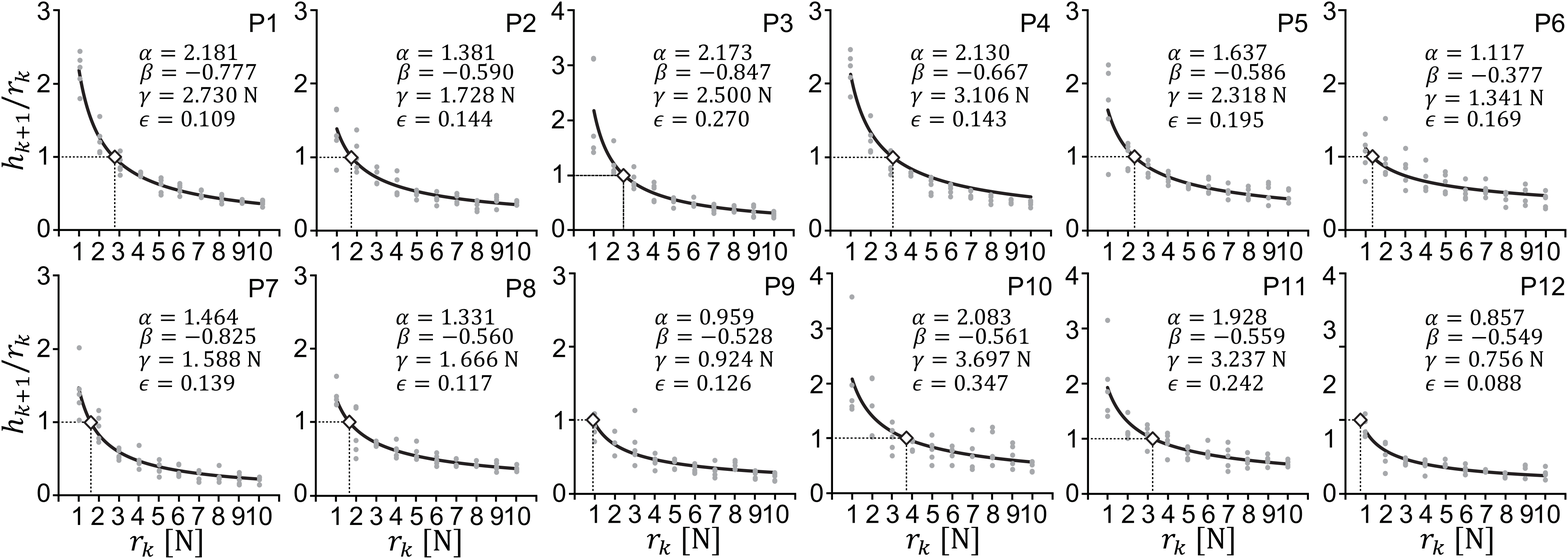}
		\end{center}
	\end{minipage}
	\begin{minipage}{0.16\hsize}
		\begin{center}
			\includegraphics[height=55mm]{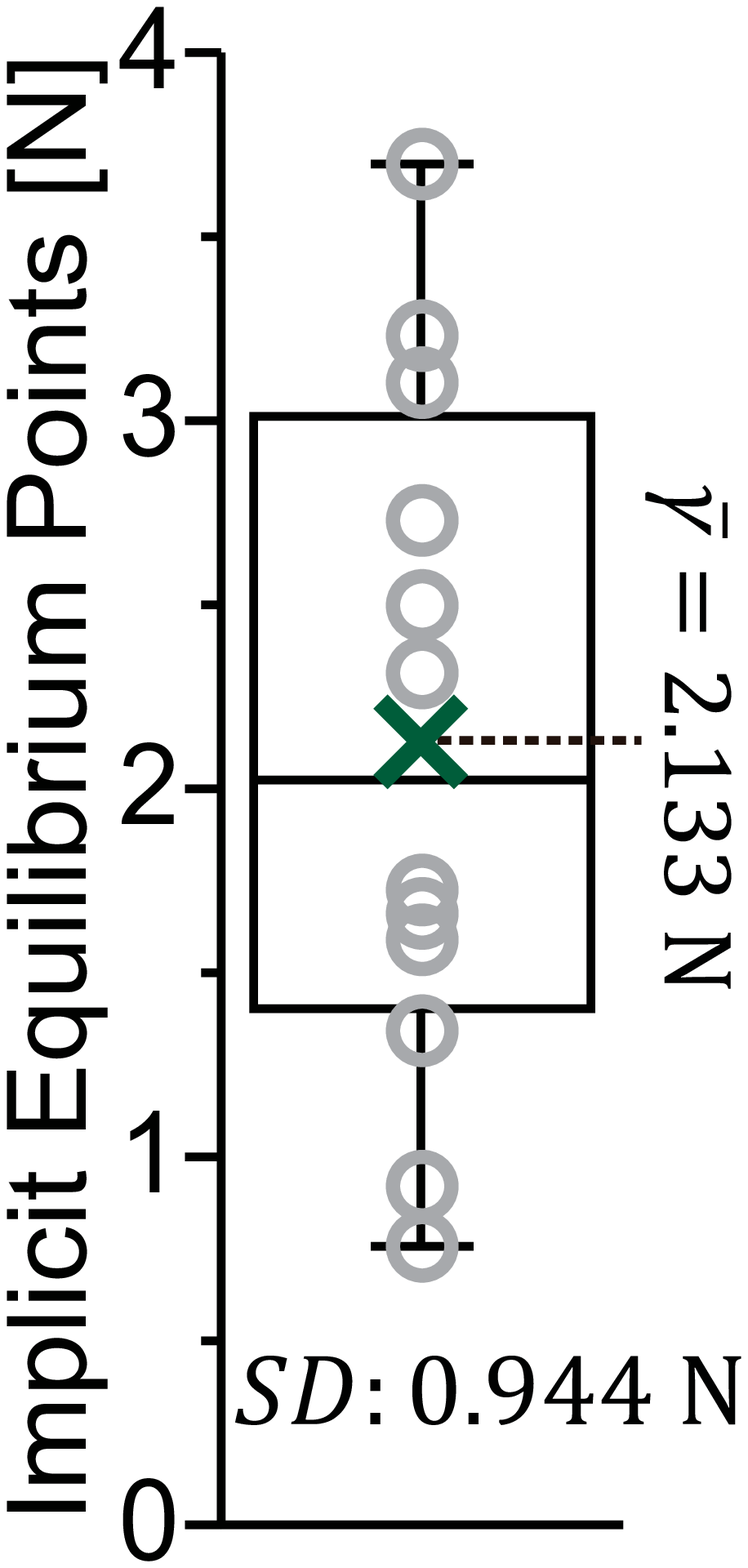}
		\end{center}
	\end{minipage}
	\begin{minipage}{0.83\hsize}
		\begin{center}
			(a)
		\end{center}
	\end{minipage}
	\begin{minipage}{0.16\hsize}
		\begin{center}
			(b)
		\end{center}
	\end{minipage}
	\begin{minipage}{0.5\hsize}
		\begin{center}
			\includegraphics[width=\hsize]{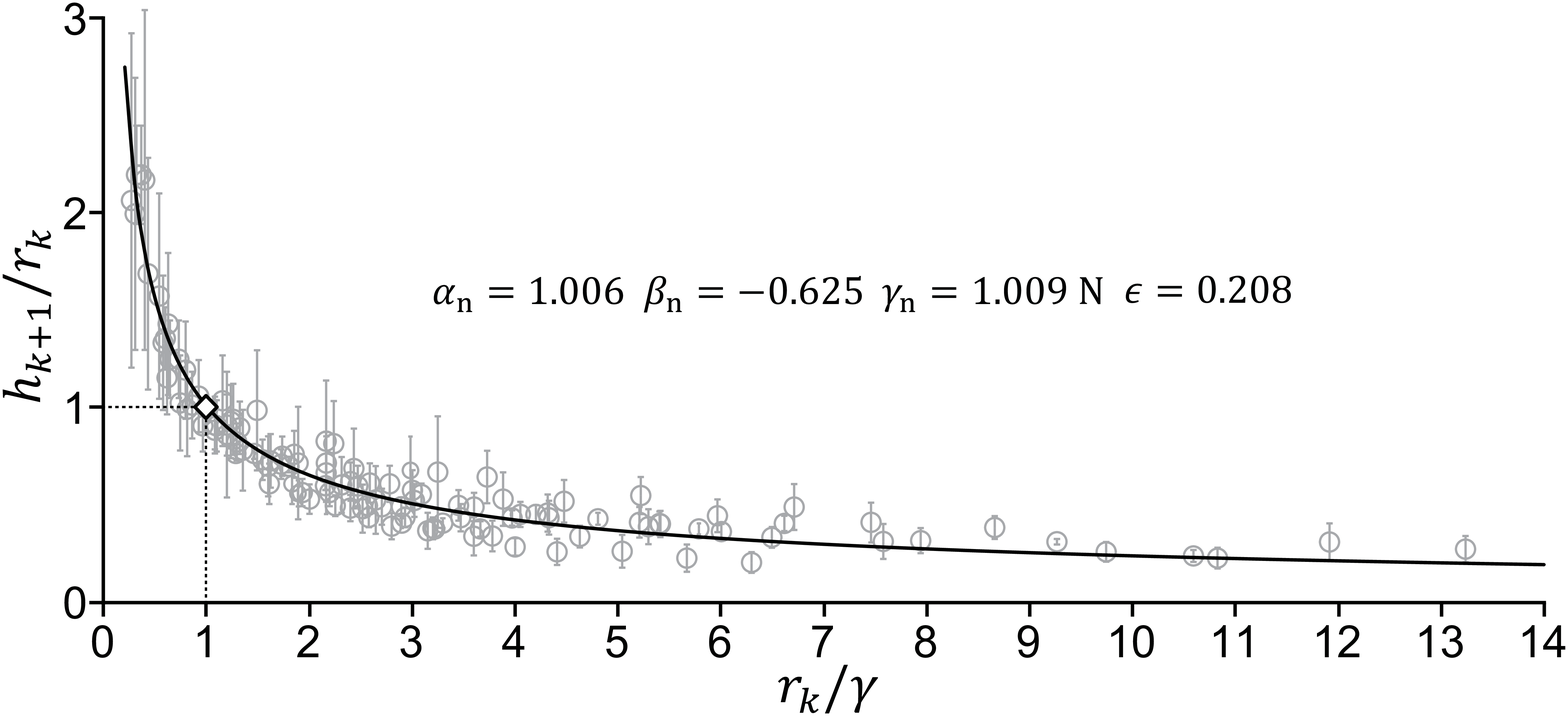}
		\end{center}
	\end{minipage}
	\begin{minipage}{0.5\hsize}
		\begin{center}
			\includegraphics[width=\hsize]{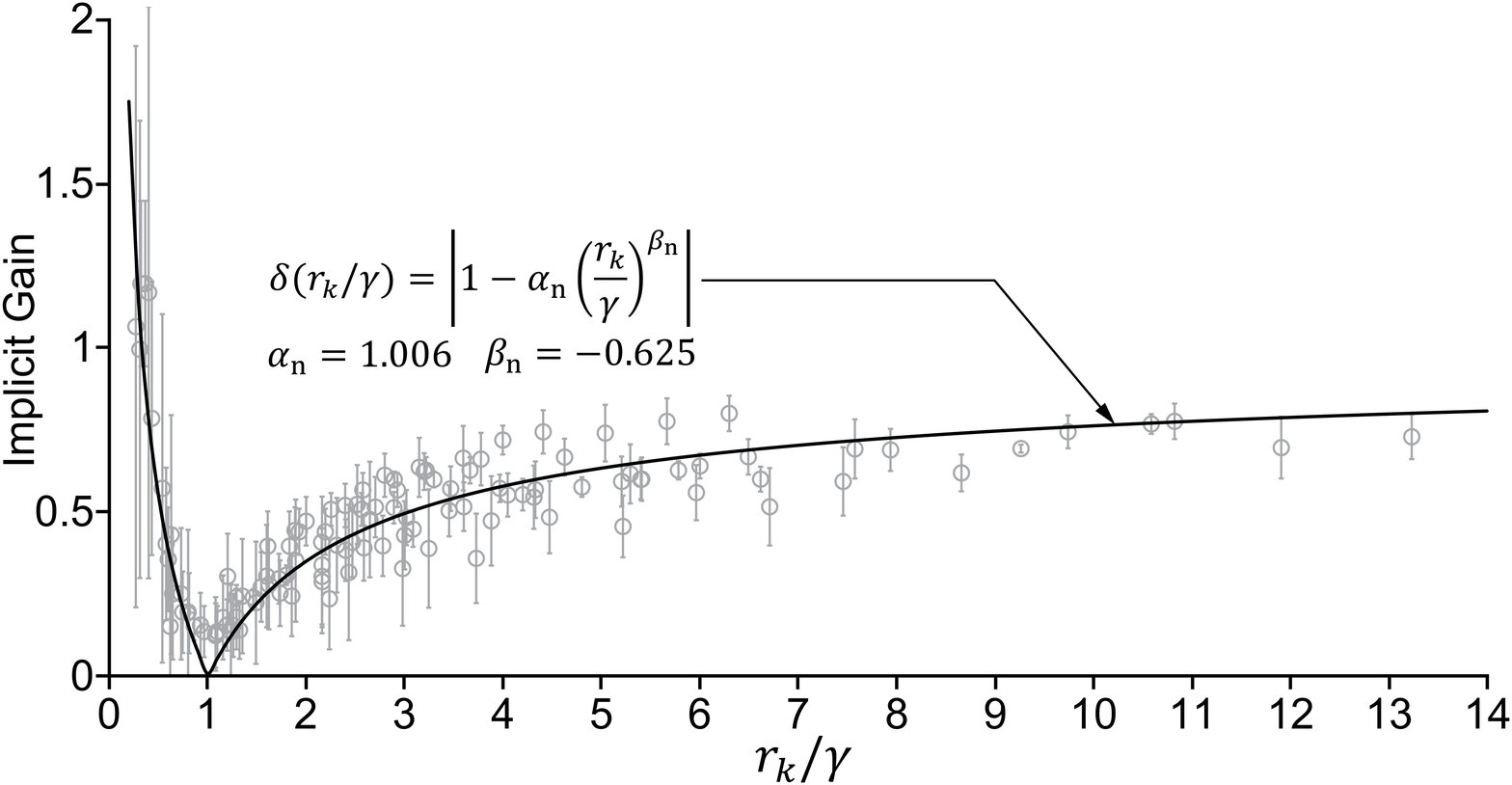}
		\end{center}
	\end{minipage}
	\begin{minipage}{0.5\hsize}
		\begin{center}
			(c)
		\end{center}
	\end{minipage}
	\begin{minipage}{0.49\hsize}
		\begin{center}
			(d)
		\end{center}
	\end{minipage}
	\begin{minipage}{0.5\hsize}
		\begin{center}
			\includegraphics[width=\hsize]{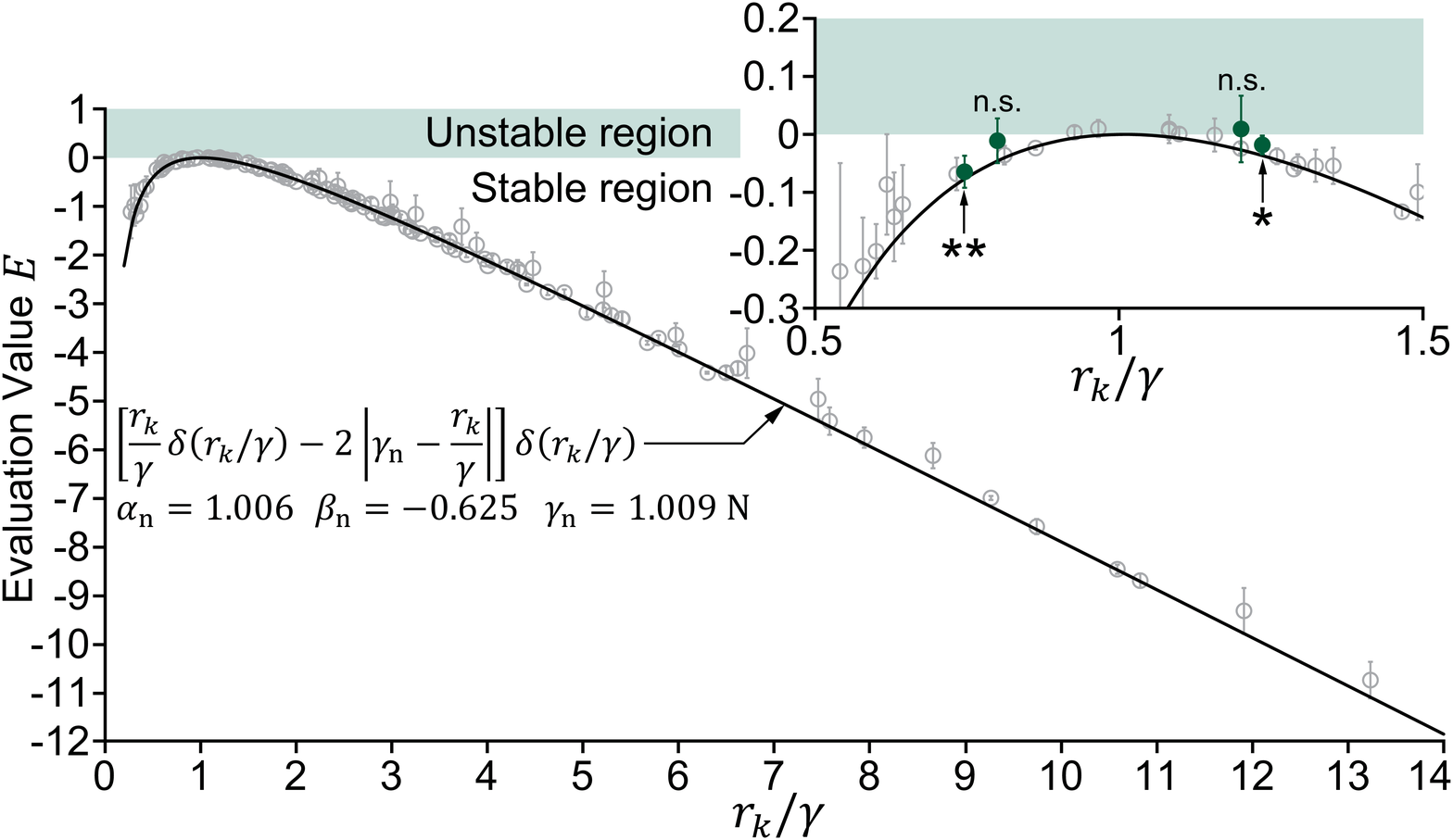}\\
			(e)
		\end{center}
	\end{minipage}
	\begin{minipage}{0.5\hsize}
	\caption{Experiment 1-A: force reproduction experiment. Experiment 1-A was conducted with twelve participants for the hand-finger interaction (Fig.~\ref{fig:ExpSys}, left). The force reproduction experiment involved fifty sets of two robot and human phases. The symbols ***, **, *, and $\mathrm{n.s.}$ represent the statistical results: $p<0.001$, $p<0.01$, $p<0.05$, and $p>0.05$, respectively.
	(a) Input-output transfer model fitting $h_{k+1}/r_k=\alpha r_k^\beta$. Each dot represents each force reproduction result, and the square represents the implicit equilibrium point.
	(b) Box plot of the implicit equilibrium points. Each circle and the cross represent each participant's point and the mean, respectively.
	(c) Normalized input-output transfer model $h_{k+1}/(r_k/\gamma)=\alpha_\mathrm{n} (r_k/\gamma)^{\beta_\mathrm{n}}$ of the twelve participants. The error bars denote the SDs.
	(d) Normalized implicit gain of the twelve participants.
	(e) Normalized asymptotic stability criteria for the twelve participants.}\label{fig:EXP1-A}
	\end{minipage}
\end{figure*}

%% file: text/section4-1.tex
\subsection{Bias Model} \label{sec:4-1}
To examine the bias model of Hypothesis 1, we conducted Experiment 1-A: force reproduction experiment under the hand-finger interaction (Fig.~\ref{fig:ExpSys} left).
The robot applied force $r_k$ to a participant in the robot phase, and the participant voluntarily applied the same force $h_{k+1}$ to the robot in the human phase.
From the interaction forces $r_k$ and $h_{k+1}$, we obtained each participant's model parameters: $\alpha$ and $\beta$ of a nonlinear input-output transfer model using nonlinear least squares (Fig.~\ref{fig:EXP1-A}(a))
\begin{align}
	\label{eq:IOT}
	\alpha r_k^\beta \coloneqq h_{k+1}/r_k,
\end{align}
where we chose the model fit for the force reproduction results.
The mean of the twelve participants' root mean square errors (RMSEs) was 0.174, and their standard deviation (SD) was 0.077.
The RMSEs $\epsilon$ used in this study were calculated by
\begin{align}
	\label{eq:}
	\epsilon \coloneqq \sqrt{\frac{1}{n}\sum_{i=1}^{n}\left(\frac{h_{(k+1)i}}{r_{ki}}-\alpha r_{ki}^{\beta}\right)^2},
\end{align}
where $i$ and $n$ denote the trial number and the total number of the trials, respectively.
The RMSEs were relatively small compared to the variations of $h_{k+1}/r_k$ between 0 and 4.
Using the model parameters $\alpha$ and $\beta$, the implicit gain and implicit equilibrium point were derived as
\begin{align}
	\label{eq:G-IEP}
	\delta(r_k)=|1-\alpha r_k^\beta|,\
	\gamma=(1/\alpha)^{1/\beta}.
\end{align}
The mean of the twelve participants' implicit equilibrium points was $\gamma=2.133\ \mathrm{N}$, and their SD was 0.944 $\mathrm{N}$ (Fig.~\ref{fig:EXP1-A}(b)).
The SD showed wide variations depending on each participant's condition.
To analyze the data of the twelve participants integrally, we normalized the interaction force applied by the robot $r_k$ as $r_k/\gamma$ using each participant's own implicit equilibrium point.
Using the normalized data, we obtained the normalized input-output transfer model (Fig.~\ref{fig:EXP1-A}(c)), where the normalized model parameters calculated by all participants' data were $\alpha_{\mathrm{n}}=1.006$ and $\beta_{\mathrm{n}}=-0.625$, and the normalized implicit equilibrium point was $\gamma_{\mathrm{n}}=1.009\ \mathrm{N}$.
The RMSE of the normalized model was 0.208.
According to \eqref{eq:G-IEP}, the normalized implicit gain was calculated using the normalized model parameters as $\delta(r_k/\gamma)=|1-1.006(r_k/\gamma)^{-0.625}|$ (Fig.~\ref{fig:EXP1-A}(d)).

%% file: text/section4-2.tex
\subsection{Stability Prediction} \label{sec:4-2}
For Hypothesis 2, we further analyzed the force reproduction results from Experiment 1-A whether the force reproduction satisfied the asymptotic stability condition \eqref{eq:AScondition}.
Using the normalized force $r_k/\gamma$ and the normalized implicit gain $\delta(r_k/\gamma)$ (Fig.~\ref{fig:EXP1-A}(d)), the evaluation values $E$ defined in \eqref{eq:Evalue} for the asymptotic stability condition were derived (Fig.~\ref{fig:EXP1-A}(e)).
The asymptotic stability condition was significantly satisfied at $r_k/\gamma=0.746$ (P6, $p<0.01$) and $r_k/\gamma=1.260$ (P11, $p<0.05$).
In contrast, the condition was not significantly satisfied at $r_k/\gamma=0.800$ (P3, $p>0.05$) and $r_k/\gamma=1.201$ (P8, $p>0.05$).
Thus, stable and unstable regions existed with respect to the normalized force $r_k/\gamma$.
Subsequently, we assumed that the mean of the normalized force, which satisfies the asymptotic stability condition, and the normalized force, which does not satisfy the asymptotic stability condition, is the boundary between the stable and unstable regions.
Then, we derived the unstable region for the normalized force as
\begin{align}
	\label{eq:}
	0.773< r_k/\gamma < 1.231,
\end{align}
where the boundaries $0.773$ and $1.231$ are the mean of $r_k/\gamma=0.746$ (P6, $p<0.01$) and $r_k/\gamma=0.800$ (P3, $p>0.05$) and the mean of $r_k/\gamma=1.260$ (P11, $p<0.05$) and $r_k/\gamma=1.201$ (P8, $p>0.05$), respectively.
Furthermore, we calculated the unstable region for the normalized absolute error $|e_k|/\gamma$ as
\begin{align}
	|e_k|/\gamma < 0.5(|1-0.773|+|1-1.231|)=0.229.
\end{align}
Therefore, we predicted that the interaction forces $r_k$ and $h_k$ would converge toward his or her own implicit equilibrium point $\gamma$ calculated by \eqref{eq:G-IEP} if the interaction forces were in the stable region $|e_k|/\gamma \geq 0.229$ and would diverge if the forces were in the unstable region $|e_k|/\gamma < 0.229$.

%% file: text/Figure/EXP1-B.tex
\begin{figure}[t!]
	\begin{minipage}{\hsize}
		\begin{center}
			\includegraphics[width=\hsize]{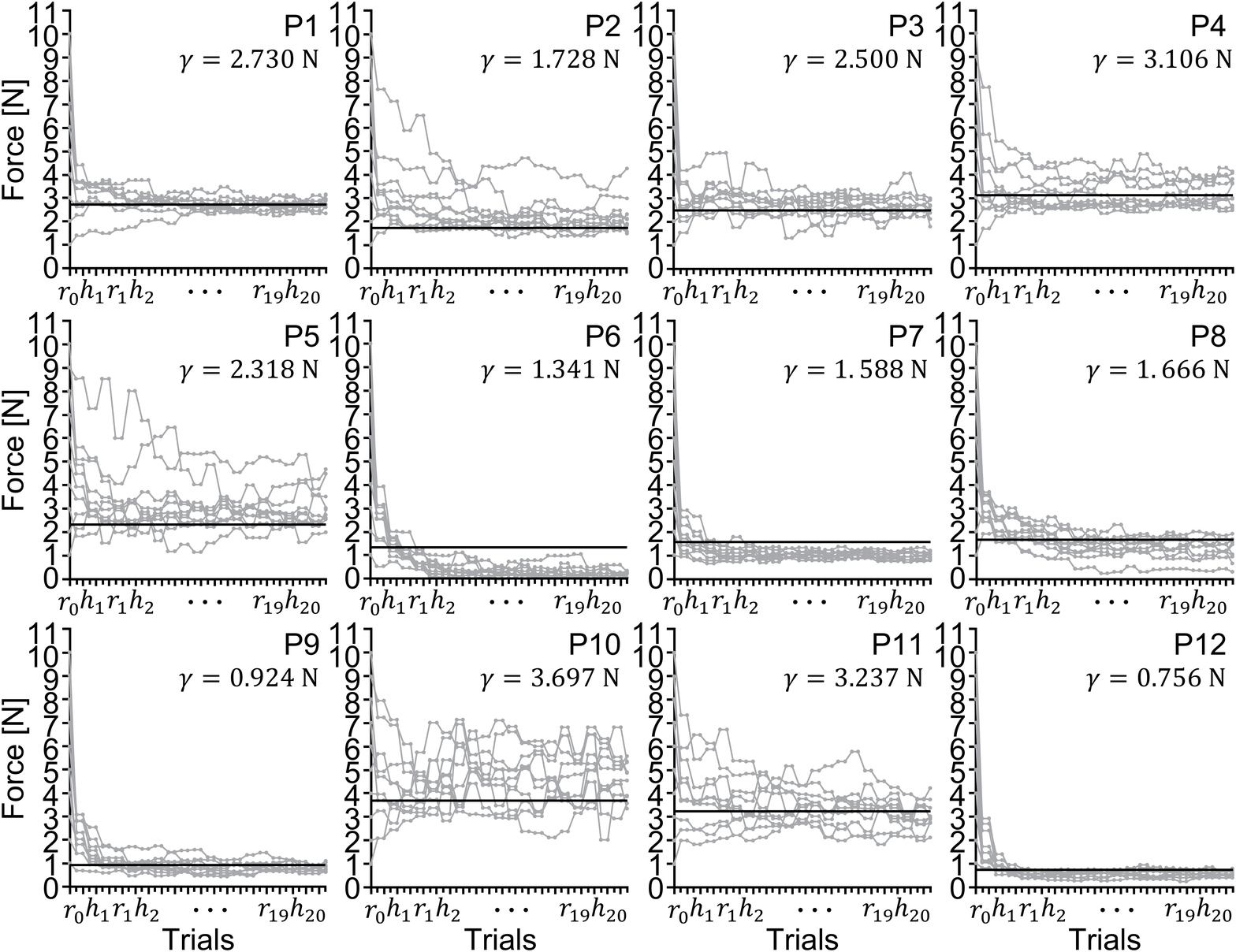}
		\end{center}
	\end{minipage}
	\begin{minipage}{\hsize}
		\begin{center}
			(a)
		\end{center}
	\end{minipage}
	\begin{minipage}{\hsize}
		\begin{center}
		\includegraphics[width=\hsize]{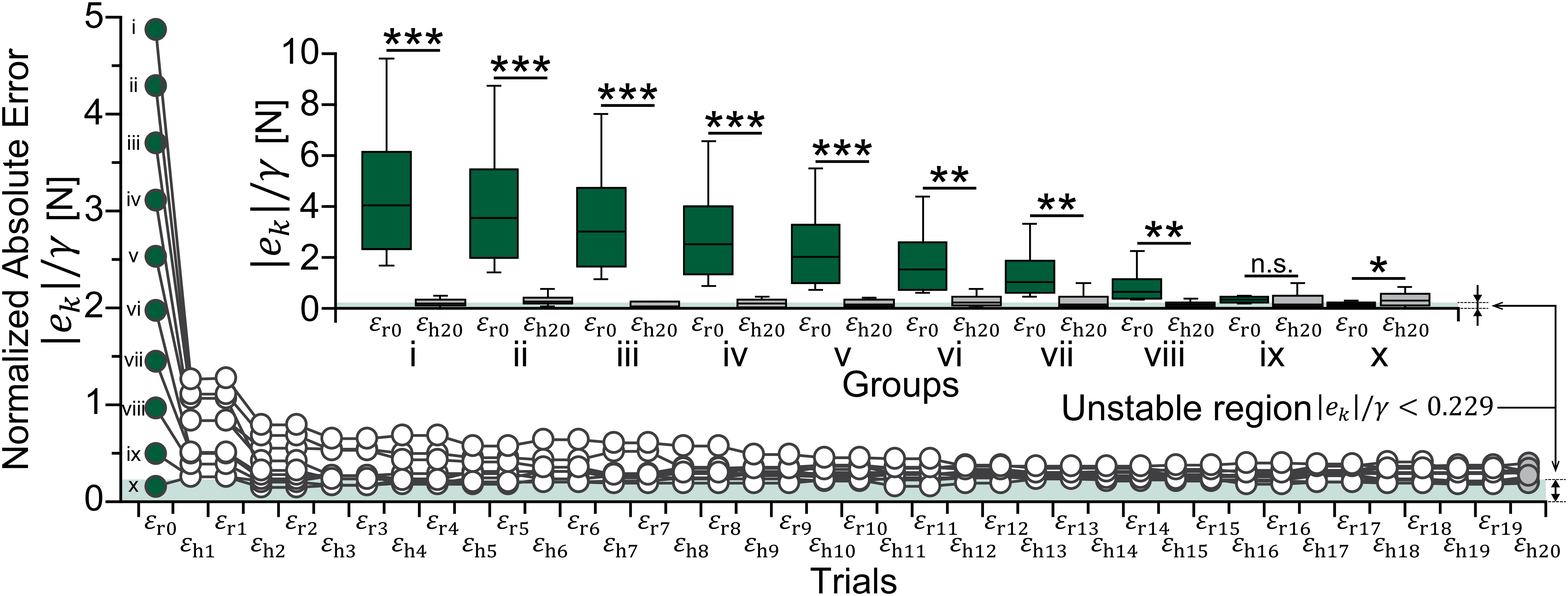}
		\end{center}
	\end{minipage}
	\begin{minipage}{\hsize}
		\begin{center}
			(b)
		\end{center}
	\end{minipage}
	\caption{Experiment 1-B: interaction experiment. Experiment 1-B was conducted with twelve participants for the hand-finger interaction (Fig.~\ref{fig:ExpSys}, left). The interaction experiment involved ten sets of forty robot and human phases (Fig.~\ref{fig:concept}). The symbols ***, **, *, and $\mathrm{n.s.}$ represent the statistical results: $p<0.001$, $p<0.01$, $p<0.05$, and $p>0.05$, respectively.
	(a) Interaction forces $h_k$ and $r_k$, which converged toward their own implicit equilibrium point $\gamma$. The points, which are represented by the lines, were calculated as Fig.~\ref{fig:EXP1-A}(a).
	(b) Means of the normalized absolute errors $\epsilon_\mathrm{rk}=|\gamma-r_k|/\gamma$ and $\epsilon_\mathrm{hk}=|\gamma-h_k|/\gamma$ from Groups i--x and the box plots of the initial normalized absolute errors $\epsilon_\mathrm{r0}$ and final normalized absolute errors $\epsilon_\mathrm{h20}$. The circles represent the means.}\label{fig:EXP1-B}
\end{figure}

%% file: text/section4-3.tex
\subsection{Convergence and Divergence of Discrete-Event Human-Robot Force Interaction} \label{sec:4-3}
Based on the prediction in Section~\ref{sec:4-2}, we conducted Experiment 1-B: interaction experiment under the hand-finger interaction (Fig.~\ref{fig:concept}, Fig.~\ref{fig:ExpSys} left).
The robot applied the same interaction force $r_k$ as the previous interaction force from a participant $h_k$ in the robot phase.
The participant voluntarily applied the same force $h_{k+1}$ to the robot in the human phase.
The robot and human phases were alternately conducted.
The same twelve participants from Experiment 1-A also took part in Experiment 1-B.
The experimental results showed that each participant's interaction forces $h_k$ and $r_k$ converged toward but not at the predicted each participant's own implicit equilibrium point $\gamma$ (Fig.~\ref{fig:EXP1-B}(a)).
To statistically evaluate the results, we classified each participant's ten interactions into Groups i--x.
Group i has the interaction with the largest initial normalized absolute error $\epsilon_{\mathrm{r}0}\coloneqq|\gamma-r_0|/\gamma$, while Group x has the interaction with the smallest normalized initial absolute error.
Accordingly, the other eight interactions were classified into Groups ii--ix.
The means of the normalized absolute errors: $\epsilon_{\mathrm{r}k}\coloneqq|\gamma-r_k|/\gamma$ and $\epsilon_{\mathrm{h}k}\coloneqq|\gamma-h_k|/\gamma$ converged toward the boundary between the stable and unstable regions $0.229$ (Fig.~\ref{fig:EXP1-B}(b)).
The final normalized absolute errors $\epsilon_\mathrm{\mathrm{h}20}$ of Groups i--viii were significantly smaller than their initial normalized absolute errors $\epsilon_{\mathrm{r}0}$ (Groups i--v: $p<0.001$, Groups vi--viii: $p<0.01$).
Conversely, the final absolute error of Group x was significantly larger than its initial absolute error ($p<0.05$).
The final normalized absolute error of Group ix was not significantly smaller than its initial normalized absolute error.

%% file: text/Figure/EXP2.tex
\begin{figure}[t!]
	\begin{minipage}{0.49\hsize}
		\begin{center}
		\includegraphics[width=\hsize]{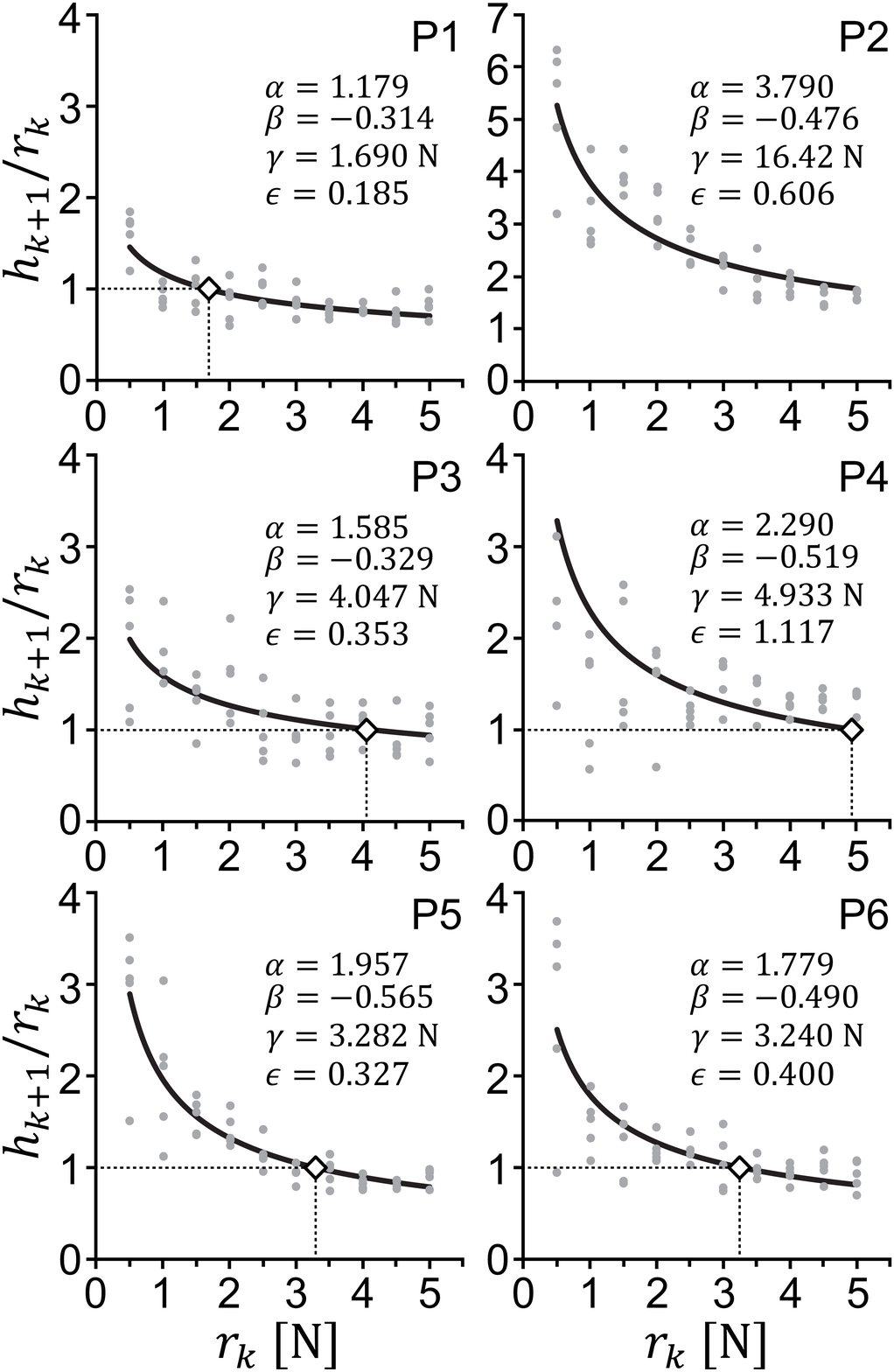}
		\end{center}
	\end{minipage}
	\begin{minipage}{0.49\hsize}
		\begin{center}
		\includegraphics[width=\hsize]{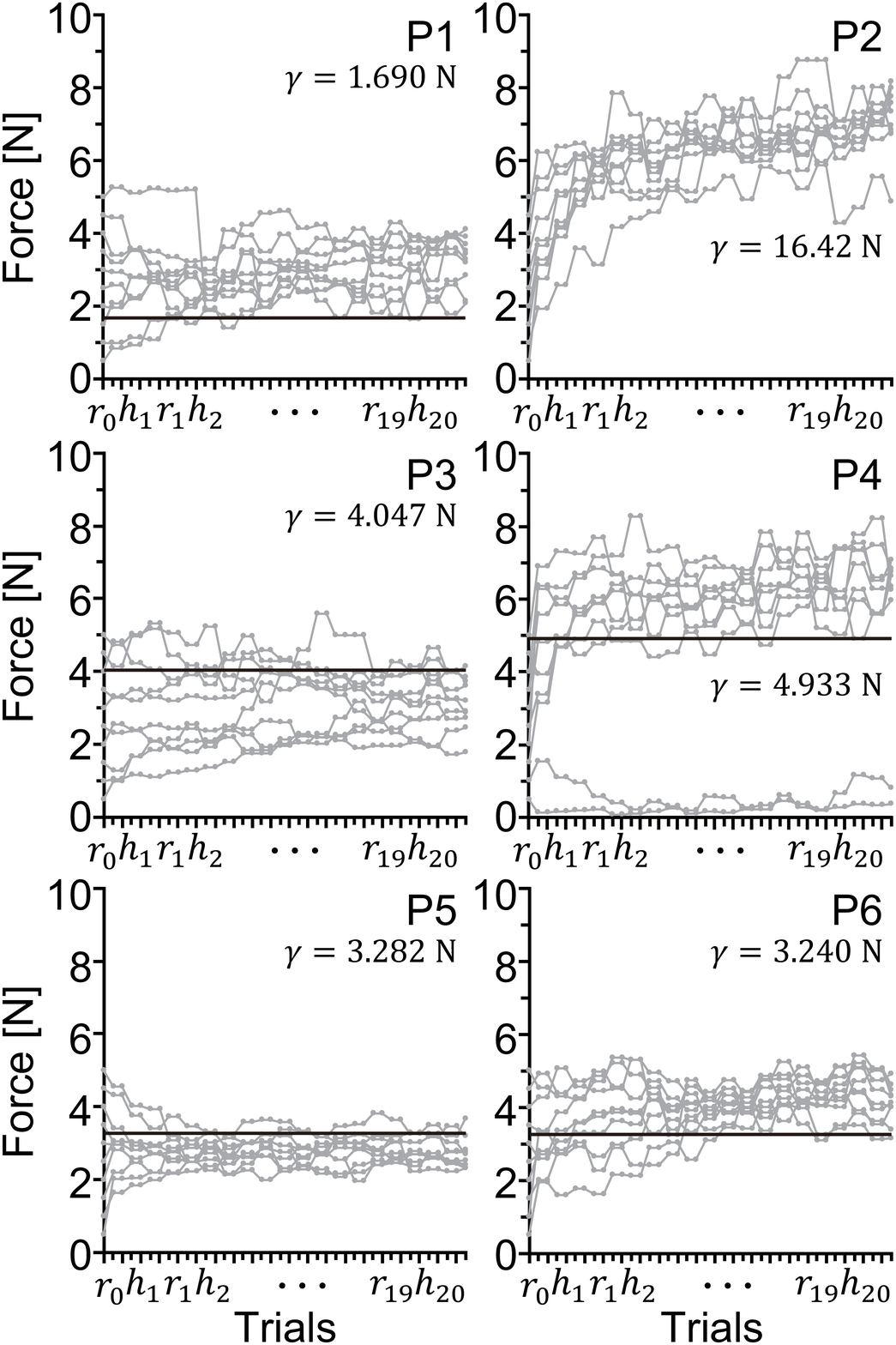}
		\end{center}
	\end{minipage}
	\begin{minipage}{0.49\hsize}
		\begin{center}
		(a)
		\end{center}
	\end{minipage}
	\begin{minipage}{0.49\hsize}
		\begin{center}
		(b)
		\end{center}
	\end{minipage}
	\begin{minipage}{\hsize}
		\begin{center}
		\includegraphics[width=\hsize]{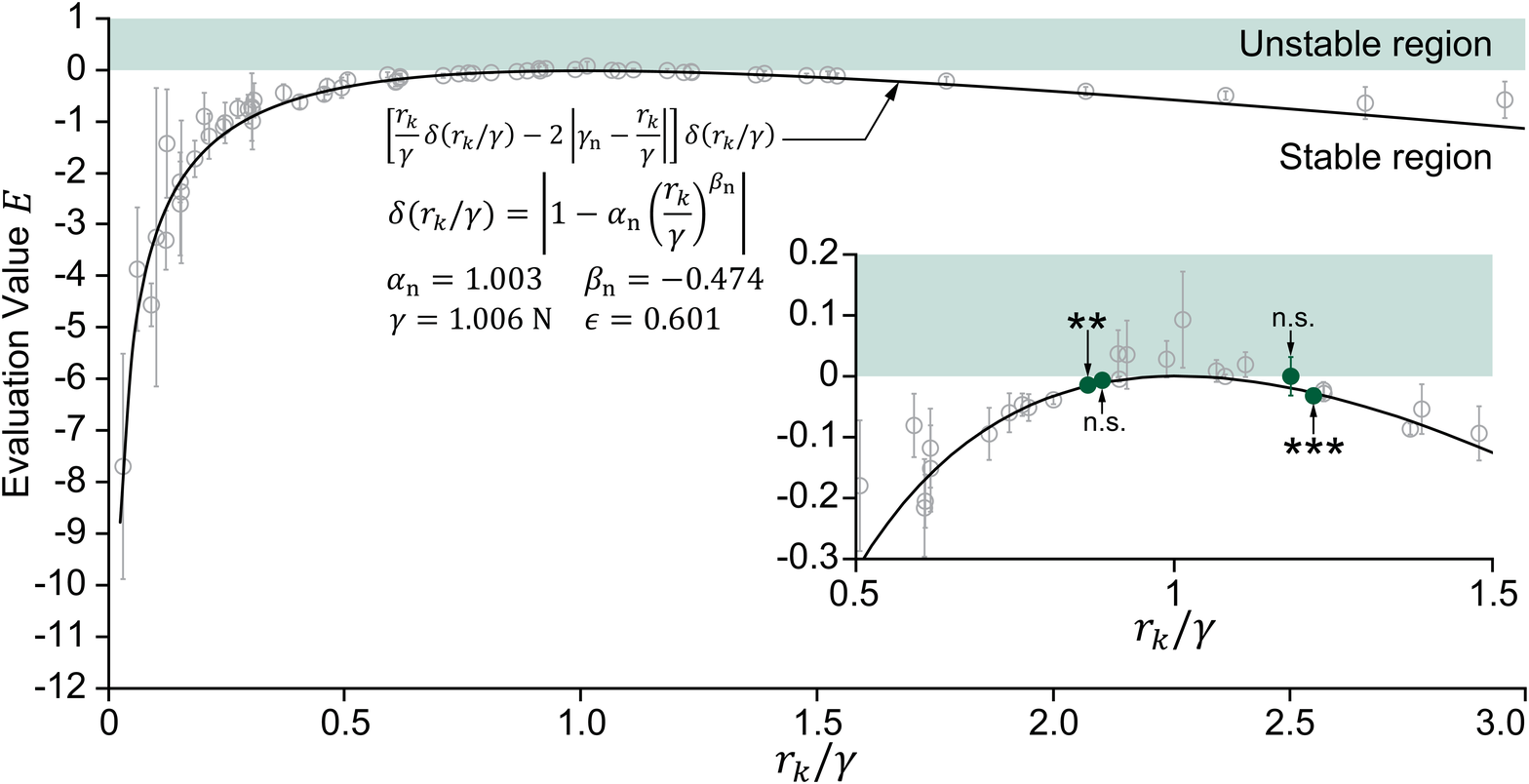}
		\end{center}
	\end{minipage}
	\begin{minipage}{\hsize}
		\begin{center}
		(c)
		\end{center}
	\end{minipage}
	\begin{minipage}{\hsize}
		\begin{center}
		\includegraphics[width=\hsize]{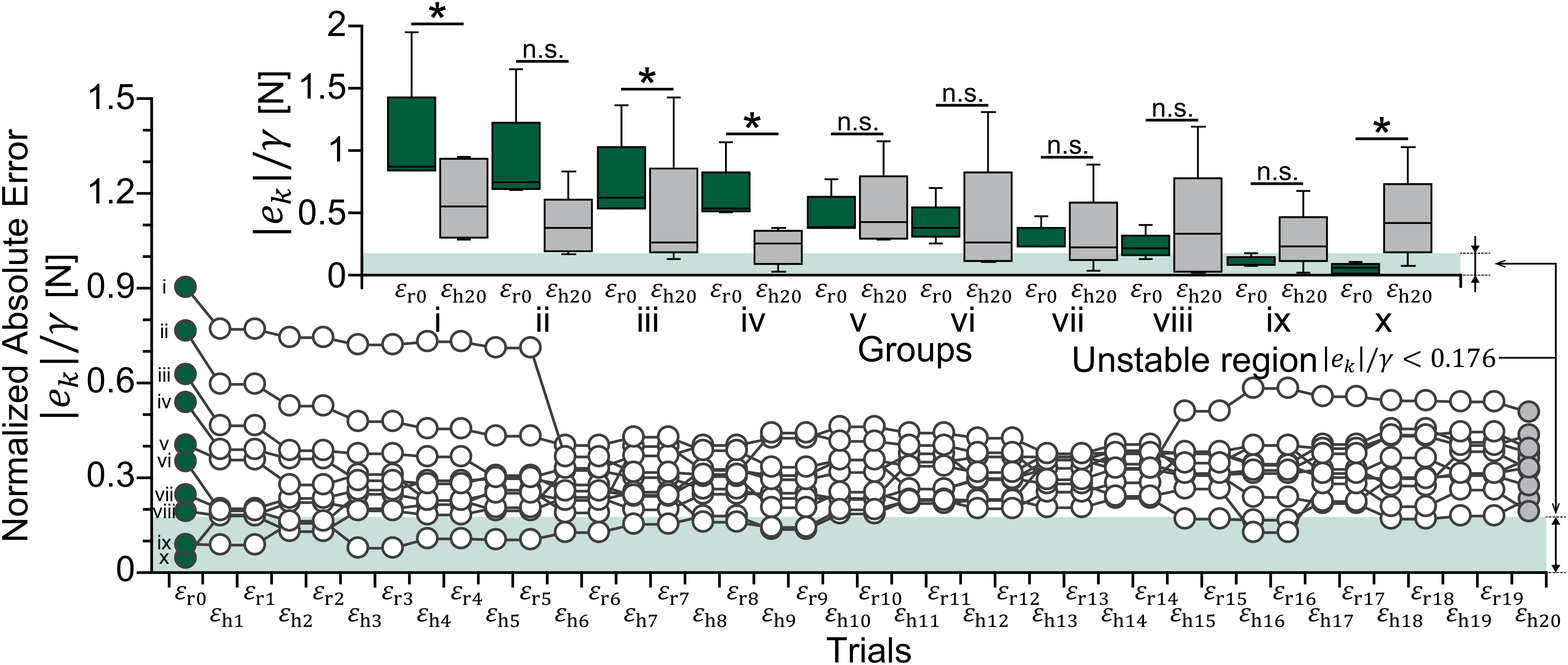}
		\end{center}
	\end{minipage}
	\begin{minipage}{\hsize}
		\begin{center}
		(d)
		\end{center}
	\end{minipage}
	\caption{Experiment 2: force reproduction (Experiment 2-A) and interaction (Experiment 2-B) experiments with six participants for the wrist interaction (Fig.~\ref{fig:ExpSys}, middle). The force reproduction experiment performed fifty sets of two robot and human phases, and the interaction experiment involved ten sets of forty robot and human phases. The symbols ***, **, *, and $\mathrm{n.s.}$ represent the statistical results: $p<0.001$, $p<0.01$, $p<0.05$, and $p>0.05$, respectively.
	(a) Input-output transfer model fitting $h_{k+1}/r_k=\alpha r_k^\beta$. Each dot represents each force reproduction result, and the square represents the implicit equilibrium point.
	(b) Interaction forces $h_k$ and $r_k$, which converged toward their own implicit equilibrium point $\gamma$. The points, which are represented by the lines, were calculated as Fig.~\ref{fig:EXP2}(a).
	(c) Normalized asymptotic stability criteria of the six participants. The error bars denote the SDs.
	(d) Means of the normalized absolute errors $\epsilon_\mathrm{rk}=|\gamma-r_k|/\gamma$ and $\epsilon_\mathrm{hk}=|\gamma-h_k|/\gamma$ from Groups i--x and the box plots of the initial normalized absolute errors $\epsilon_\mathrm{r0}$ and final normalized absolute errors $\epsilon_\mathrm{h20}$. The circles represent the means.}
	\label{fig:EXP2}
\end{figure}

%% file: text/Figure/EXP3.tex
\begin{figure}[t!]
	\begin{minipage}{0.49\hsize}
		\begin{center}
		\includegraphics[width=\hsize]{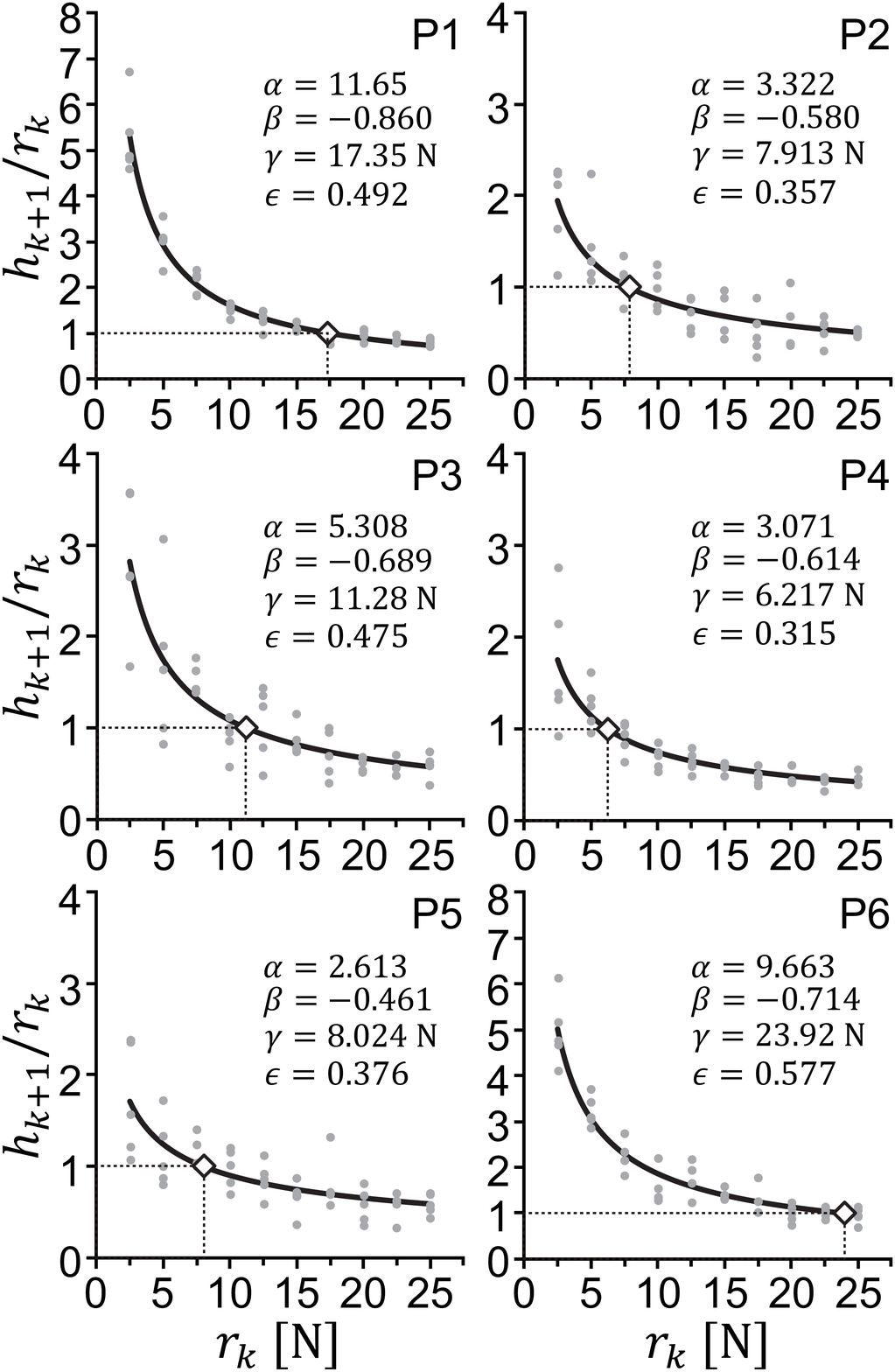}
		\end{center}
	\end{minipage}
	\begin{minipage}{0.49\hsize}
		\begin{center}
		\includegraphics[width=\hsize]{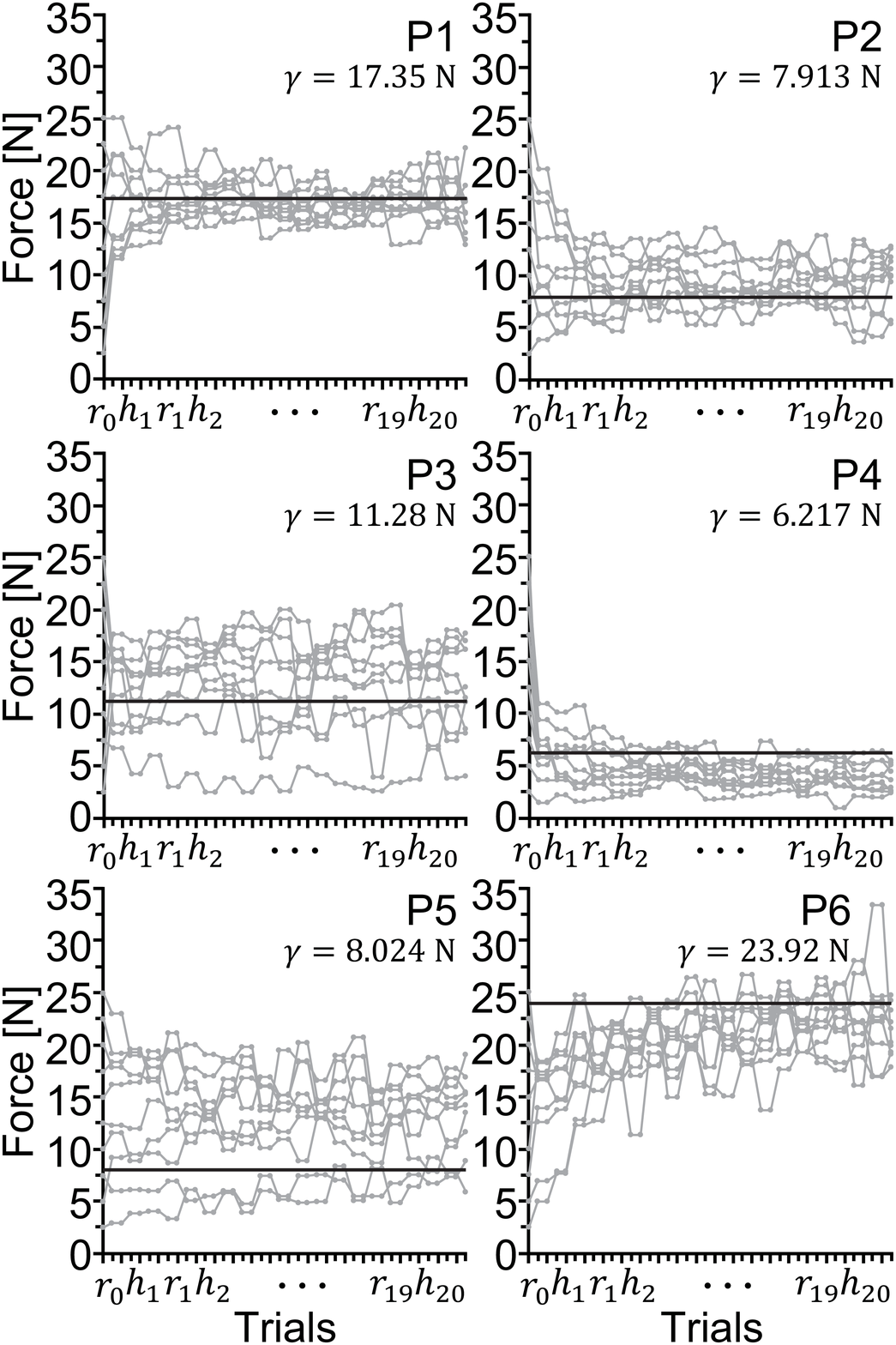}
		\end{center}
	\end{minipage}
	\begin{minipage}{0.49\hsize}
		\begin{center}
		(a)
		\end{center}
	\end{minipage}
	\begin{minipage}{0.49\hsize}
		\begin{center}
		(b)
		\end{center}
	\end{minipage}
	\begin{minipage}{\hsize}
		\begin{center}
		\includegraphics[width=\hsize]{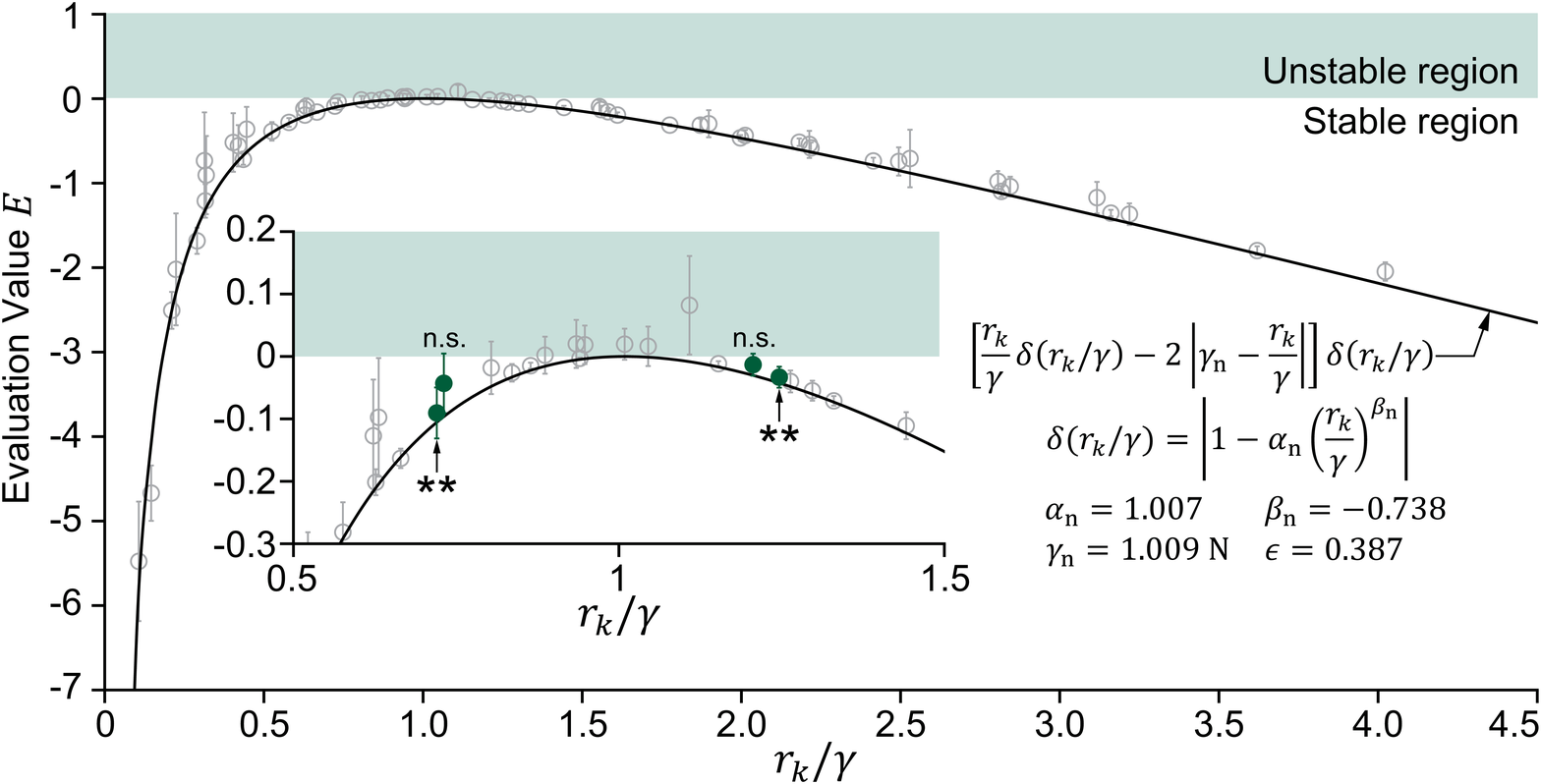}
		\end{center}
	\end{minipage}
	\begin{minipage}{\hsize}
		\begin{center}
		(c)
		\end{center}
	\end{minipage}
	\begin{minipage}{\hsize}
		\begin{center}
		\includegraphics[width=\hsize]{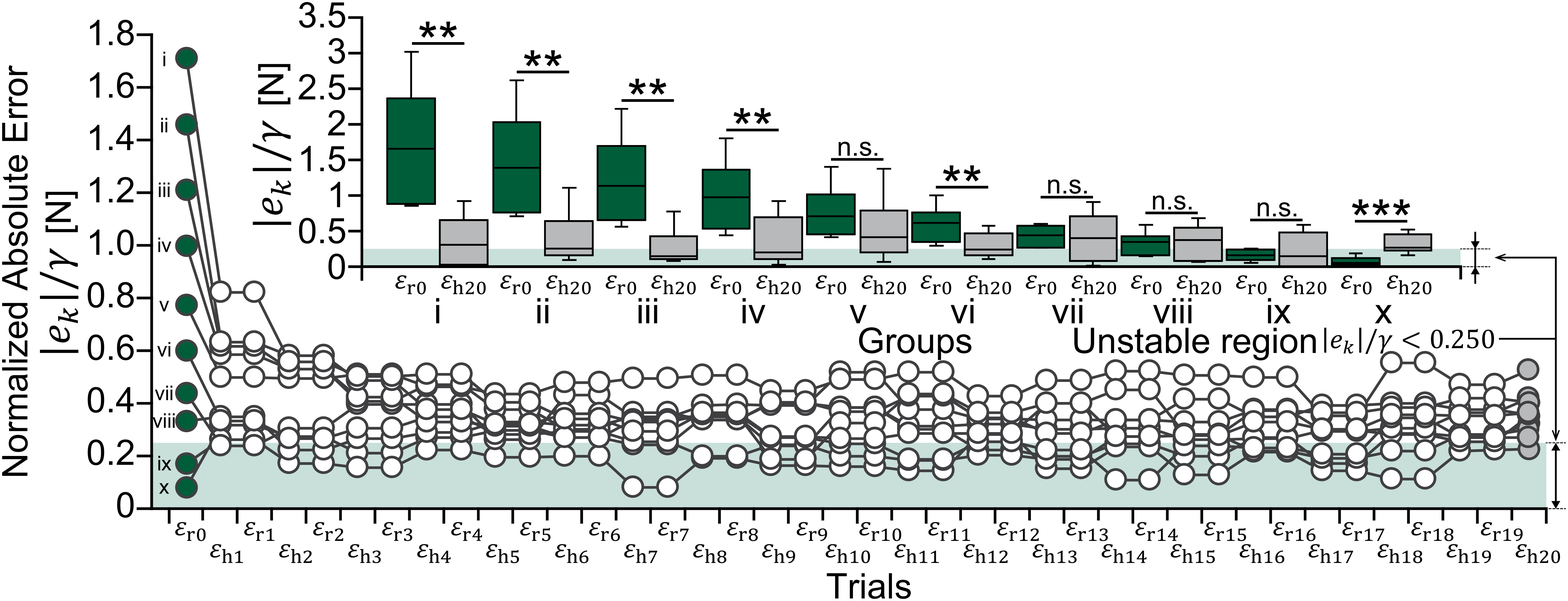}
		\end{center}
	\end{minipage}
	\begin{minipage}{\hsize}
		\begin{center}
		(d)
		\end{center}
	\end{minipage}
	\caption{Experiment 3: force reproduction (Experiment 3-A) and interaction (Experiment 3-B) experiments with six participants for the foot interaction (Fig.~\ref{fig:ExpSys}, right). The force reproduction experiment performed fifty sets of two robot and human phases, and the interaction experiment involved ten sets of forty robot and human phases. The symbols ***, **, *, and $\mathrm{n.s.}$ represent the statistical results: $p<0.001$, $p<0.01$, $p<0.05$, and $p>0.05$, respectively.
	(a) Input-output transfer model fitting $h_{k+1}/r_k=\alpha r_k^\beta$. Each dot represents each force reproduction result, and the square represents the implicit equilibrium point.
	(b) Interaction forces $h_k$ and $r_k$, which converged toward their own implicit equilibrium point $\gamma$. The points, which are represented by the lines, were calculated as Fig.~\ref{fig:EXP3}(a).
	(c) Normalized asymptotic stability criteria of the six participants. The error bars denote the SDs.
	(d) Means of the normalized absolute errors $\epsilon_\mathrm{rk}=|\gamma-r_k|/\gamma$ and $\epsilon_\mathrm{hk}=|\gamma-h_k|/\gamma$ from Groups i--x and the box plots of the initial normalized absolute errors $\epsilon_\mathrm{r0}$ and final normalized absolute errors $\epsilon_\mathrm{h20}$. The circles represent the means.}
	\label{fig:EXP3}
\end{figure}

%% file: text/section4-4.tex
\subsection{Generalizability} \label{sec:4-4}
To verify the generalizability of the hypotheses, we conducted four experiments under two different conditions: wrist interaction and foot interaction (Fig.~\ref{fig:ExpSys} middle and right).

For the wrist interaction (Fig.~\ref{fig:ExpSys} middle), we conducted Experiments 2-A and 2-B, which were force reproduction experiment (Fig.~\ref{fig:EXP2}(a)) and interaction experiment (Fig.~\ref{fig:EXP2}(b)), respectively.
Six participants took part in both the experiments.
According to Experiment 2-A, the mean and SD of the six RMSEs in the model fitting were 0.498 and 0.332, respectively.
Additionally, the mean and SD of the implicit equilibrium points were 5.603 N and 5.408 N, respectively.
For the asymptotic stability analysis, the evaluation values \eqref{eq:AScondition} were calculated (Fig.~\ref{fig:EXP2}(c)).
The asymptotic stability condition was significantly satisfied at $r_k/\gamma=0.811$ (P3, $p<0.01$) and $r_k/\gamma=1.219$ (P5, $p<0.001$), and the condition was not significantly satisfied at $r_k/\gamma=0.887$ (P1, $p>0.05$) and $r_k/\gamma=1.183$ (P1, $p>0.05$).
In the same manner as the derivation of the unstable region in Section~\ref{sec:4-2}, we derived the unstable region of the wrist interaction with respect to the normalized force as $0.849 < r_k/\gamma < 1.201$, and the unstable region for the normalized absolute error was calculated as $|e_k|/\gamma < 0.176$.
According to Experiment 2-B, we classified each participant's ten interactions into Groups i--x in the same manner as the groups in the hand-finger interaction (Fig.~\ref{fig:EXP2}(d)).
The means of the normalized absolute errors converged toward above the boundary $0.176$ between the stable and unstable regions.
The final normalized absolute errors $\epsilon_\mathrm{\mathrm{h}20}$ of Groups i, iii, and iv were significantly smaller than their initial normalized absolute errors $\epsilon_\mathrm{\mathrm{r}0}$ ($p<0.05$).
In contrast, the final normalized absolute error of Group x was significantly larger than its initial normalized absolute error ($p<0.05$).
The final normalized absolute errors of Groups ii, v, vi, vii, and viii were not significantly smaller than their initial normalized absolute errors, and the final normalized absolute error of Group ix was not significantly larger than its initial normalized absolute error.
%

For the foot interaction (Fig.~\ref{fig:ExpSys} right), we conducted Experiments 3-A and 3-B, which were force reproduction experiments (Fig.~\ref{fig:EXP3}(a)) and interaction experiments (Fig.~\ref{fig:EXP3}(b)), respectively.
Six participants took part in both the experiments.
In Experiment 3-A, the mean and SD of the six RMSEs in the model fitting were 0.432 and 0.099, respectively.
Additionally, the mean and SD of the implicit equilibrium points were 12.450 N and 6.870 N, respectively.
For the asymptotic stability analysis, the evaluation values \eqref{eq:AScondition} were calculated (Fig.~\ref{fig:EXP3}(c)).
The asymptotic stability condition was significantly satisfied at $r_k/\gamma=0.721$ (P1, $p<0.01$) and $r_k/\gamma=1.246$ (P5, $p<0.01$), and the condition was not significantly satisfied at $r_k/\gamma=0.732$ (P6, $p>0.05$) and $r_k/\gamma=1.206$ (P4, $p>0.05$).
In the same manner as the derivation of the unstable region in Section~\ref{sec:4-2}, we derived the unstable region of the foot interaction with respect to the normalized force as $0.727< r_k/\gamma < 1.226$, and the unstable region for the normalized absolute error was calculated as $|e_k|/\gamma < 0.250$.
According to Experiment 3-B, we classified each participant's ten interactions into Groups i--x in the same manner as the groups in the hand-finger interaction (Fig.~\ref{fig:EXP3}(d)).
The means of the normalized absolute errors converged toward above the boundary $0.250$ between the stable and unstable regions.
The final normalized absolute errors $\epsilon_\mathrm{\mathrm{h}20}$ of Groups i, ii, iii, iv, and vi were significantly smaller than their initial normalized absolute errors $\epsilon_\mathrm{\mathrm{r}0}$ ($p<0.01$).
In contrast, the final normalized absolute error of Group x was significantly larger than its initial normalized absolute error ($p<0.001$).
The final normalized absolute errors of Groups v, vii, and viii were not significantly smaller than their initial normalized absolute errors, and the final normalized absolute error of Group ix was not significantly larger than its initial normalized absolute error.

%% file: text/section4-5.tex
\subsection{Comparison of Three Interactions} \label{sec:4-5}
We compared the hand-finger, wrist, and foot interactions to investigate whether the wider (narrower) unstable region, which was predicted by the force reproduction experiments and Hypothesis~2, results in the greater (less) variance of the interaction convergence.

First, according to the asymptotic stability criteria based on the force reproduction experiments in Experiments 1-A, 2-A, and 3-A (Fig.~\ref{fig:EXP1-A}(e), \ref{fig:EXP2}(c), and \ref{fig:EXP3}(c), respectively), we counted the number of the convergent and divergent evaluation values $E$ in the range $0.5 \leq r_k/\gamma \leq 1.5$.
The total number of the convergent evaluation values and that of the divergent evaluation values were 118 and 17 for the hand-finger interaction, 110 and 30 for the wrist interaction, and 102 and 23 for the foot interaction.
Moreover, we calculated divergence rates at each range of the normalized force $r_k/\gamma$ as the lines in Fig.~\ref{fig:Comparison}(a) with
\begin{align}
	\label{eq:}
	&\mathrm{Divergence\ rate}\ [\%] \coloneqq 100\times\notag\\
	&\frac{\mathrm{Num.\ of\ divergent\ }E}{\mathrm{Num.\ of\ convergent\ }E + \mathrm{Num.\ of\ divergent\ }E}.
\end{align}
The sum of the divergence rates was 129 \% for the hand-finger interaction, 197 \% for the wrist interaction, and 190 \% for the foot interaction.
Hence, the probabilities that interaction force would converge around an implicit equilibrium point were predicted to be different among the three interactions in the order Wrist Interaction (Experiment 2-A) $<$ Foot Interaction (Experiment 3-A) $<$ Hand-Finger Interaction (Experiment 1-A).

Next, according to the interaction experiments in Experiments 1-B, 2-B, and 3-B (Figs.~\ref{fig:EXP1-B}(a), \ref{fig:EXP2}(b), and \ref{fig:EXP3}(b), respectively), we counted the distribution of the normalized final interaction force $h_{20}/\gamma$ (Fig.~\ref{fig:Comparison}(b)) and calculated asymptotic convergence rates with
\begin{align}
	\label{eq:}
	&\mathrm{Asym.\ conv.\ rate}\ [\%] \coloneqq 100\times\notag\\
	&\frac{\mathrm{Num.\ of\ trials\ whose}\ h_{20}/\gamma\ \mathrm{was\ between\ 0.75\ and\ 1.25}}{\mathrm{Num.\ of\ Trials}}.
\end{align}
The asymptotic convergence rate was 59 \% for the hand-finger interaction, 28 \% for the wrist interaction, and 45 \% for the foot interaction.
Hence, the rates that the interaction force converged around an implicit equilibrium point were different among the three interactions in the order Wrist Interaction (Experiment 2-B) $<$ Foot Interaction (Experiment 3-B) $<$ Hand-Finger Interaction (Experiment 1-B).
This rate order was consistent with the predicted probability order based on the results of the force reproduction experiments.

%% file: text/Figure/Comparison.tex
\begin{figure}[t!]
	\begin{minipage}{\hsize}
		\begin{center}
			\includegraphics[width=\hsize]{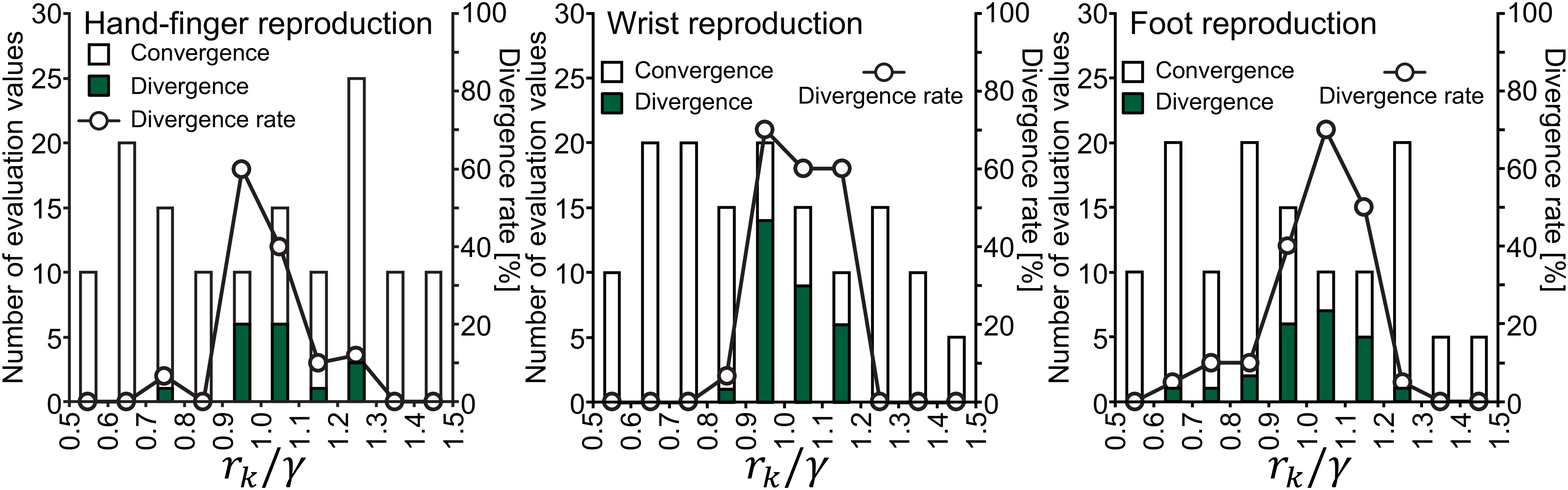}
		\end{center}
	\end{minipage}
	\begin{minipage}{\hsize}
		\begin{center}
			(a)
		\end{center}
	\end{minipage}
	\begin{minipage}{\hsize}
		\begin{center}
			\includegraphics[width=\hsize]{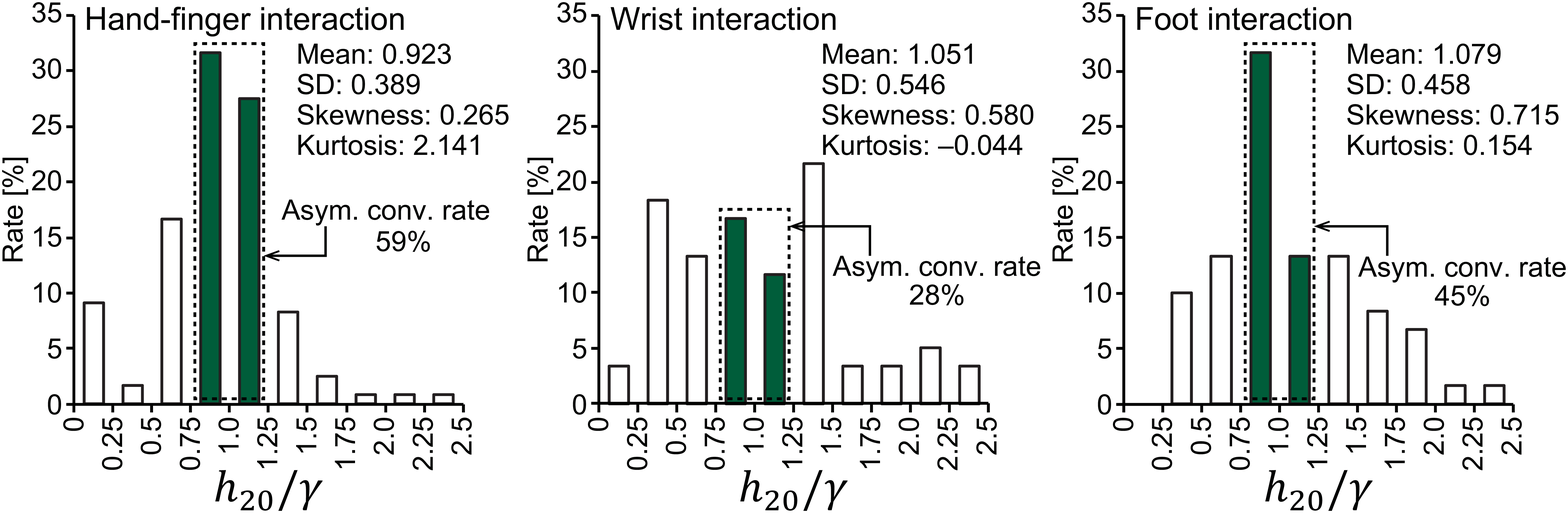}
		\end{center}
	\end{minipage}
	\begin{minipage}{\hsize}
		\begin{center}
			(b)
		\end{center}
	\end{minipage}
	\caption{Comparison of the hand-finger, wrist, and foot interactions.
	(a) Rates of the divergent force reproductions in Experiments 1-A, 2-A, and 3-A. The convergent and divergent samples were counted according to the evaluation values in $0.5<r_k/\gamma<1.5$ from the normalized asymptotic stability criteria (Figs.~\ref{fig:EXP1-A}(e), \ref{fig:EXP2}(c), and \ref{fig:EXP3}(c)).
	(b) Distribution of the normalized final interaction force $h_{20}/\gamma$ in Experiments 1-B, 2-B, and 3-B (Figs.~\ref{fig:EXP1-B}(a), \ref{fig:EXP2}(b), and \ref{fig:EXP3}(b)).}\label{fig:Comparison}
\end{figure}

%% file: text/section5.tex
\section{Discussion} \label{sec:5}
\input{text/section5-1}

\input{text/section5-2}

\input{text/Figure/IFC}

\input{text/section5-3}

\input{text/section5-4}

%% file: text/section5-1.tex
\subsection{Hypothesis 1} \label{sec:5-1}
The bias model \eqref{eq:Bias} is composed of two variables: implicit gain $\delta(r_k)$ and implicit equilibrium point $\gamma$ to over-reproduce in the low-force level and under-reproduce in the high-force level.
The implicit gain is used to express the amplitude of the bias, and the implicit equilibrium point is used to express the sign of the bias as
\begin{align}
	\label{eq:}
	\mathrm{sgn}(\gamma-r_k)=\left\{
	\begin{array}{cl}
		1&\mathrm{if}\ \gamma > r_k\\
		0&\mathrm{if}\ \gamma = r_k\\
		-1&\mathrm{if}\ \gamma < r_k\\
	\end{array}
	\right. .
\end{align}
The sign function reflects that force reproduction by a human has a systematic error depending on reproducing force level \cite{walsh2011,Onneweer2013,Onneweer2016,takagi2016}.
We expected that the boundary of the sign inversion corresponds to the equilibrium point of the discrete-event human-robot force interaction, and we employed $\mathrm{sgn}(\gamma -r_k)$ into the bias model.
Because $\mathrm{sgn}(\gamma -r_k )$ only determines the sign, we added the implicit gain $\delta(r_k)$ as the amplitude.
Consequently, we hypothesized the bias model \eqref{eq:Bias}.

According to Experiment 1-A for the hand-finger interaction (Fig.~\ref{fig:EXP1-A}(a)), the bias on force reproduction satisfactorily obeyed the model \eqref{eq:Bias} with the small RMSEs.
In the bias model, the twelve implicit equilibrium points showed the wide variations and SD (Fig.~\ref{fig:EXP1-A}(b)), which might be because the implicit equilibrium point $\gamma$ is an individual characteristic that differs among individuals and depends on a musculoskeletal system and accuracy of self-generated force perception \cite{valles2013,Bays2006}.
In contrast, the implicit gain $\delta(r_k)$ appeared to be a common function (Fig.~\ref{fig:EXP1-A}(d)).
By normalizing the implicit equilibrium points, we attenuated the individual differences of the bias model and combined participants' models into a single integrated model (Fig.~\ref{fig:EXP1-A}(c)).

In addition to Experiment 1-A for the hand-finger interaction, we further examined the bias model with Experiments 2-A and 3-A for the wrist and foot interactions.
The means of the RMSEs in the model fitting for the hand-finger, wrist, and foot interactions were 0.174, 0.432, and 0.498, respectively (Figs.~\ref{fig:EXP1-A}(a), \ref{fig:EXP2}(a), and \ref{fig:EXP3}(a), respectively).
Therefore, Hypothesis 1 holds for the three different tasks using the three different body parts: hand, wrist, and foot, and the hypothesis may have a potential capacity holding for other tasks.

The accuracy of the bias model $U$ depends on a type of a movement and perception for a task.
The fitting RMSE of the foot interaction (mean: 0.432) was significantly larger ($p<0.001$) than that of the hand-finger interaction (mean: 0.174).
This was because of an increase in the implicit equilibrium points, as the implicit equilibrium point of the foot interaction (mean: 12.450 N) was significantly larger ($p<0.01$) than that of the hand-finger interaction (mean: 2.133 N).
Thus, the model accuracy did not deteriorate for the foot interaction compared to the hand-finger interaction.
In contrast, although the fitting RMSE of the wrist interaction (mean: 0.498) was significantly larger ($p<0.05$) than that of the hand-finger interaction, the implicit equilibrium point of the wrist interaction (mean: 5.603 N) and that of the hand-finger interaction were not significantly different ($p>0.05$).
Thus, the model accuracy might deteriorate for the wrist interaction.

%% file: text/section5-2.tex
\subsection{Hypothesis 2} \label{sec:5-2}
We evaluated whether the asymptotic stability condition \eqref{eq:AScondition} was satisfied by the force reproduction in Experiments 1-A, 2-A, and 3-A (Figs. \ref{fig:EXP1-A}(a), \ref{fig:EXP2}(a), and \ref{fig:EXP3}(a), respectively).
As a result, the asymptotic stability condition was significantly satisfied, except for the area surrounding an implicit equilibrium point (Figs. \ref{fig:EXP1-A}(e), \ref{fig:EXP2}(c), and \ref{fig:EXP3}(c), respectively).
Thus, Hypothesis 2 and results of Experiments 1-A, 2-A, and 3-A predicted that the interaction force would converge toward his or her own implicit equilibrium point $\gamma$ and diverge near the point in the discrete-event human-robot force interaction.
According to the stability criteria, we further calculated the unstable regions: $|e_k|/\gamma<0.229$, $|e_k|/\gamma<0.176$, and $|e_k|/\gamma<0.250$ for the hand-finger, wrist, and foot interactions, respectively.
To confirm the prediction, Experiments 1-B, 2-B, and 3-B: hand-finger, wrist, and foot interaction experiments were conducted (Figs.~\ref{fig:EXP1-B}(a), \ref{fig:EXP2}(b), and \ref{fig:EXP3}(b), respectively).
Consistent with the prediction, the interaction force converged toward his or her own implicit equilibrium point and diverged near the point.
The normalized absolute errors $\epsilon_{\mathrm{r}k}$ and $\epsilon_{\mathrm{h}k}$ decreased if the error was in the stable region and increased if the error was in the unstable region (Figs. \ref{fig:EXP1-B}(b), \ref{fig:EXP2}(d), and \ref{fig:EXP3}(d), respectively).
The consistency between the prediction and results supported the bias model in Hypothesis~1 and the asymptotic stability condition in Hypothesis~2, which further indicates that the bias asymptotically stabilizes the implicit equilibrium point of the voluntarily marginally stable discrete-event human-robot force interaction far from the point and destabilizes the point near the point.
Moreover, although the accuracy of the prediction depended on the task, Hypothesis 2 was verified under the three different interactions.
This implies that Hypothesis~2 may also hold for other tasks with other body parts.
%

The interaction force, which converged toward his or her own implicit equilibrium point and diverged near the point, converged around the boundary between the stable and unstable regions.
This was because the unstable force reproduction in the unstable region around the implicit equilibrium point might cause the steady-state errors of the normalized absolute errors in the three experiments (Figs.~\ref{fig:EXP1-B}(b), \ref{fig:EXP2}(d), and \ref{fig:EXP3}(d), respectively).
In the results of Experiments 1-B, 2-B, and 3-B, the initial absolute errors $\epsilon_{\mathrm{r}0}$ of Group x, which were in the unstable region, significantly increased at the final absolute errors $\epsilon_{\mathrm{h}20}$ due to the unstable force reproduction.
Moreover, in the comparison of the three interactions in Subsection~\ref{sec:4-5}, the asymptotic convergence rate decreased (Fig.~\ref{fig:Comparison}(b)) as the divergence rate increased (Fig.~\ref{fig:Comparison}(a)).
This implies that the wider (narrower) unstable region results in the greater (less) variance of the interaction convergence corresponding to the steady-state errors.
The previous study \cite{takagi2016} investigated the convergent discrete-event human-human interaction caused by the bias, which is similar to our results.
We mathematically formulated the bias and mechanism of the convergence with the discrete-event human-robot interaction and further found the divergent interaction around the implicit equilibrium point.
In addition to \cite{walsh2011,Onneweer2013,Onneweer2016,takagi2016}, our results also reconfirmed that the force escalation in \cite{Shergill2003} might be convergence of small initial force toward a higher implicit equilibrium point.

%% file: text/Figure/IFC.tex
\begin{figure}[t!]
	\begin{minipage}{0.64\hsize}
		\begin{center}
			\includegraphics[width=\hsize]{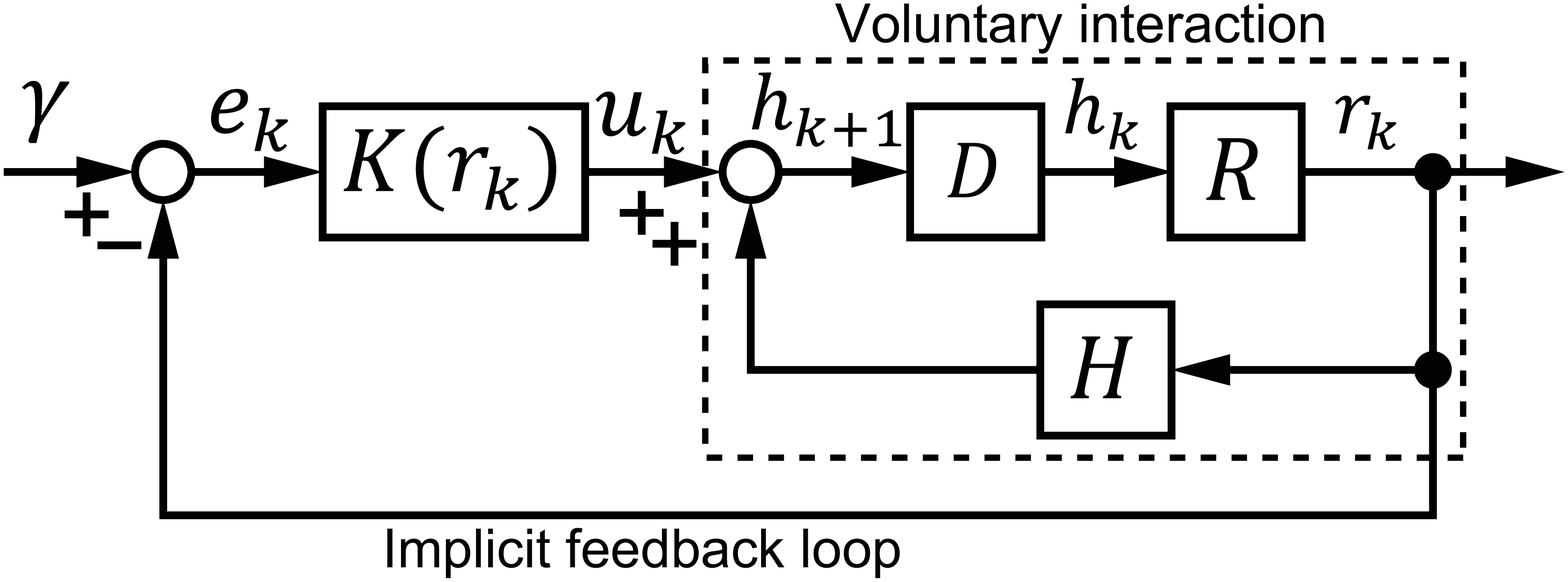}
		\end{center}
	\end{minipage}
	\begin{minipage}{0.35\hsize}
		\begin{center}
			\includegraphics[width=\hsize]{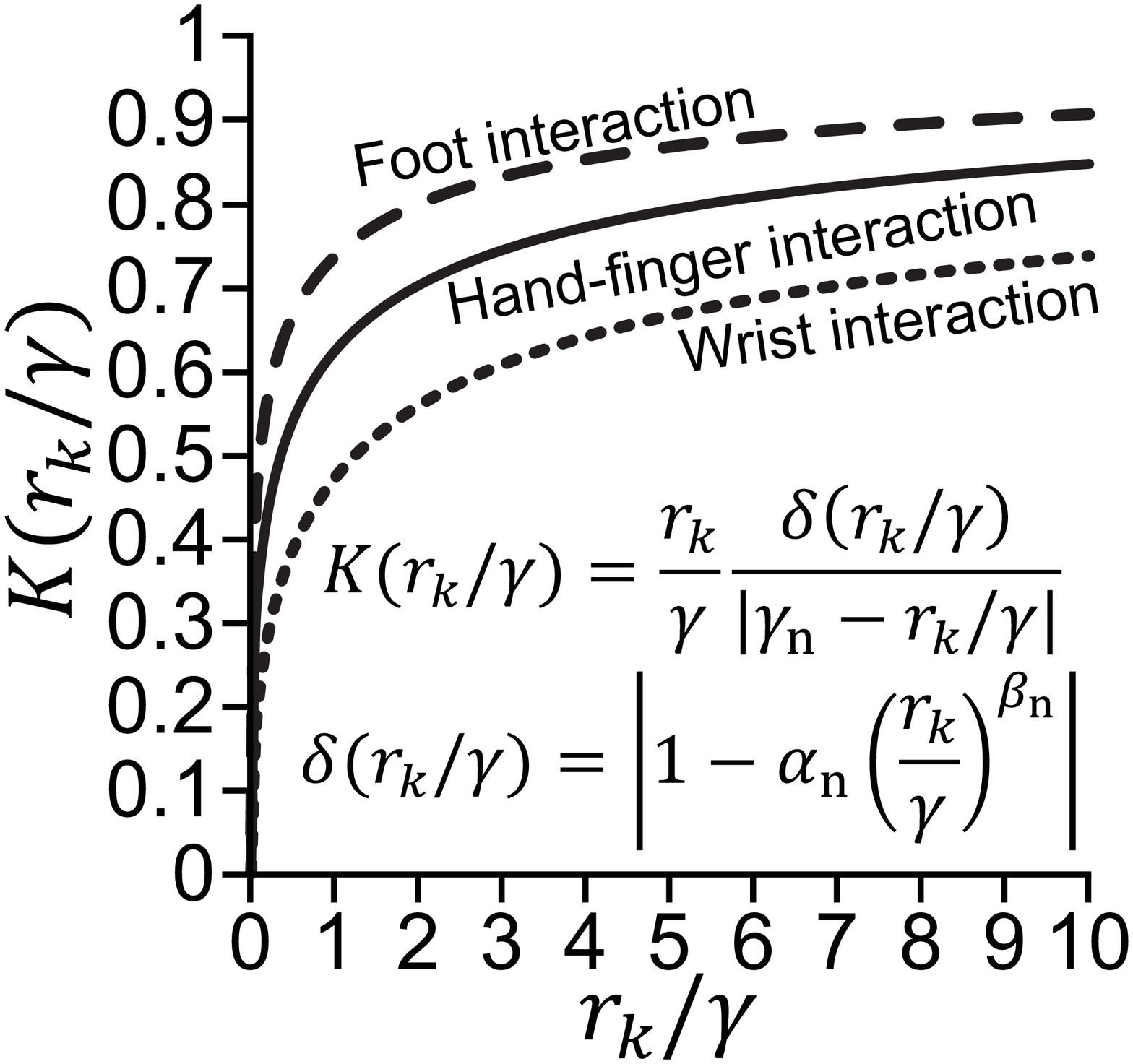}
		\end{center}
	\end{minipage}
	\begin{minipage}{0.64\hsize}
		\begin{center}
			(a)
		\end{center}
	\end{minipage}
	\begin{minipage}{0.35\hsize}
		\begin{center}
			(b)
		\end{center}
	\end{minipage}
	\caption{Human's implicit feedback control.
	(a) Block diagram of the general discrete-event human-robot force interaction, where the implicit input $u_k$ works to equalize $r_k$ with $\gamma$.
	(b) Variable gain $K(r_k/\gamma)$ of the implicit feedback control. The three variable gains were calculated using the normalized model parameters $\alpha_\mathrm{n}$ and $\beta_\mathrm{n}$ of the hand-finger, wrist, and foot interactions (Figs.~\ref{fig:EXP1-A}(c), \ref{fig:EXP2}(c), and \ref{fig:EXP3}(c)).}\label{fig:IFC}
\end{figure}

%% file: text/section5-3.tex
\subsection{Human's Implicit Feedback Control} \label{sec:5-3}
We discuss how the bias $U(r_k)$ affects general discrete-event human-robot force interaction.
Consider the following general discrete-event human-robot force interaction:
\begin{align}
	\label{eq:generalDEHRFI}
	\left\{
	\begin{array}{l}
		h_{k+1}=Hr_k + u_k(r_k)\\
		r_k=Rh_k
	\end{array}
	\right.,\ k=0,\ 1,\ 2,\ 3,\ \ldots\ ,
\end{align}
which can be voluntarily asymptotically stable or unstable, unlike the model \eqref{eq:DEHRFI}.
The variables $H$ and $R$ are the positive voluntary gains of a human and robot, respectively, and $u_k (r_k)$ denotes the implicit input.
The implicit input is derived from \eqref{eq:Bias} as
\begin{align}
	\label{eq:II}
	u_k(r_k)=U(r_k)r_k=K(r_k)[\gamma-r_k],\
	K(r_k)=\frac{\delta(r_k)}{|\gamma-r_k|}r_k,
\end{align}
which functions as a variable-gain $K(r_k)$ feedback controller that attenuates the error between the implicit equilibrium point $\gamma$ and interaction force $r_k$.
The block diagram of the general discrete-event human-robot force interaction illustrates that the implicit feedback control by a human adjusts the interaction to equalize $r_k$ with $\gamma$ (Fig.~\ref{fig:IFC}(a)).
The variable gain $K(r_k)$ changes between 0 and 1 with respect to the interaction force $r_k$ (Fig.~\ref{fig:IFC}(b)).

The bias $U(r_k)$ generally functions as implicit feedback control and provides a positive stabilization effect on the general discrete-event human-robot force interaction.
The error $e_k$ is defined by the implicit equilibrium point $\gamma$ and the interaction force of the robot $r_k$ as \eqref{eq:error:DEF}.
The error dynamics of the general interaction \eqref{eq:generalDEHRFI} are
\begin{align}
	\label{eq:}
	e_{k+1}=R(H-K)e_k.
\end{align}
The variable gain satisfies
\begin{align}
	\label{eq:}
	K(0)&=0\ \mathrm{if}\ r_k=0\\
	\lim_{r_k\to\infty}K(r_k)&=1\ \because \beta<0
\end{align}
based on \eqref{eq:G-IEP} and \eqref{eq:II}.
The zero convergences of P6 in the hand-finger interaction (Fig.~\ref{fig:EXP1-B}(a)) and P4 in the wrist interaction (Fig.~\ref{fig:EXP2}(b)) might be caused by the zero gain $K(0)=0$.
The transition $R(H-K)$ satisfies
\begin{align}
	\label{eq:RHK}
	\begin{array}{cl}
		|R(H-K)|\leq RH&\mathrm{if}\ K \leq H\\
		|R(H-K)|\leq R&\mathrm{if}\ H < K
	\end{array}
	\because 0 \leq K \leq 1,\ \forall r_k .
\end{align}
Although the stability of the implicit equilibrium point $\gamma$ of the general interaction depends on the voluntary gains $H$ and $R$ because the asymptotic stability condition is $|R(H-K)|<1$ if $e_k\neq0$, the variable gain $K$ reduces $RH$ or $R$ as $|R(H-K)|$, which is the positive stabilization effect.

%% file: text/section5-4.tex
\subsection{Limitations} \label{sec:5-4}
There are three limitations in this study.
First, we considered the discrete-event human-robot force interaction (Fig.~\ref{fig:concept}) and did not consider a continuous human-robot force interaction.
Second, the interactions (Fig.~\ref{fig:ExpSys}) were one-dimensional interactions, and the generalizability of the results to a multi-dimensional interactions was not investigated.
Third, four participants in Experiment 2 (wrist) also took part in Experiment 1 (hand-finger), and the participants in Experiments 2 and 3 (foot) were the same.
Nevertheless, participating in the hand-finger (Experiment 1), wrist (Experiment 2), and/or foot (Experiment 3) interactions might not change the their bias and results of interactions because we did not provide their experimental results and the participants were unaware of their involuntary behavior.

%% file: text/conclusion.tex
\section{Conclusion} \label{sec:6}
We set up the two hypotheses for the involuntary behavior in the discrete-event human-robot interaction.
Hypothesis~1 provides the mathematical bias model, and Hypothesis~2 provides the asymptotic stability condition for the voluntarily marginally stable interaction in consideration of the bias.
Hypothesis~1 was supported by the force reproduction experiments.
Hypothesis~2 and the results of the force reproduction experiments predicted that the interaction would converge toward his or her own implicit equilibrium point and diverge around the point.
The interaction experiments examined and showed that the interaction force significantly converged toward the point, and the steady-state errors occurred owing to the divergence, which supported Hypothesis~2.
The hypotheses were supported by the three different interactions using different body parts: a hand finger, wrist, and foot, which implies the generalizability of the hypotheses.
Moreover, the bias could provide the stabilization effect on the general discrete-event human-robot interaction as a human's implicit feedback control.

%% file: text/thanks.tex
\section*{Acknowledgment}
We would like to thank Dr. Hirokazu Yanagihara, Professor of Hiroshima University, and Dr. Keisuke Fukui, Associate Professor of Hiroshima University, for lending their expertise on the statistical data analysis.

%% file: text/references.bbl

%% file: text/biography.tex
\begin{IEEEbiography}
[{\includegraphics[width=1in,height=1.25in,clip,keepaspectratio]{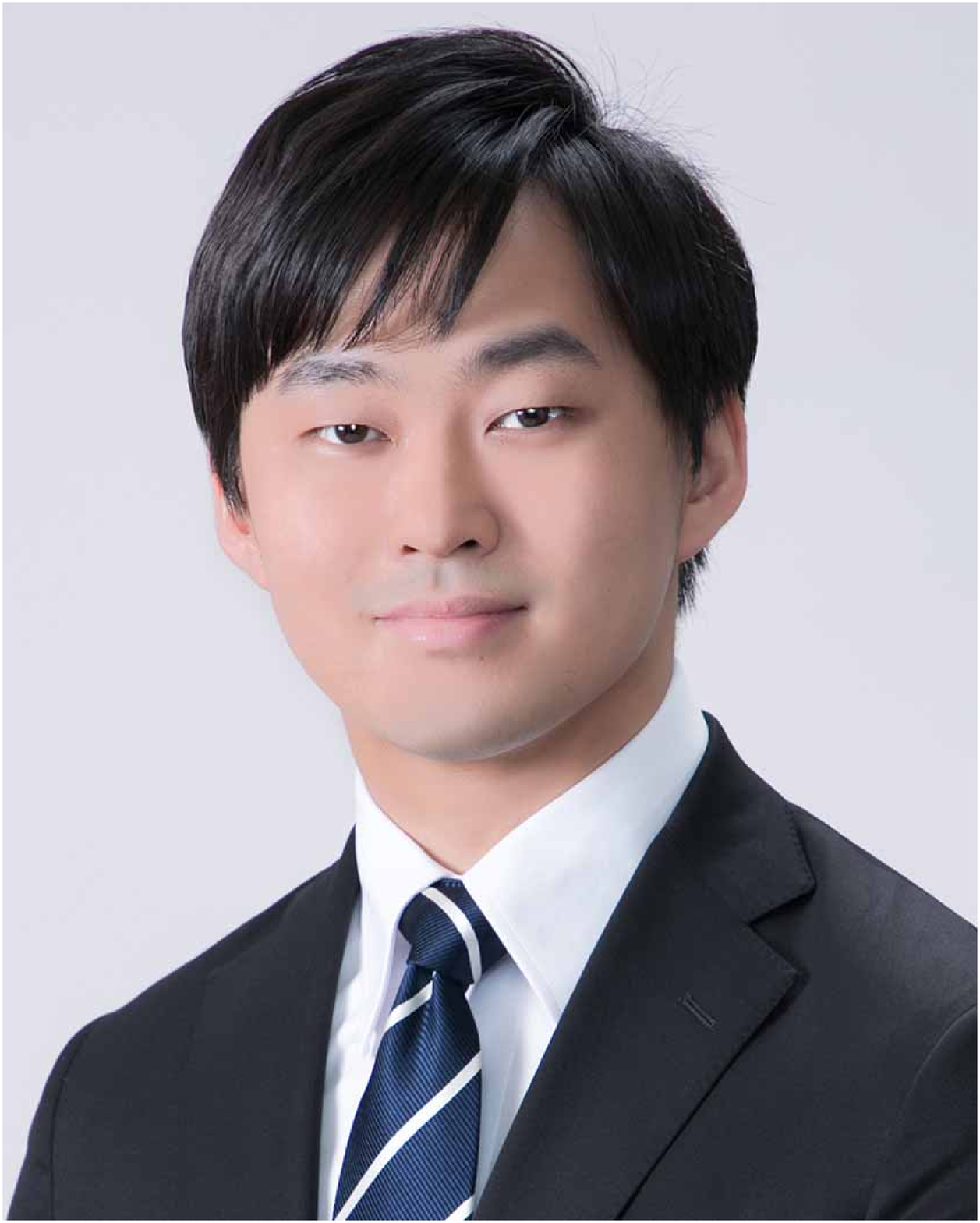}}]
{Hisayoshi Muramatsu} (Member, IEEE)
received the B.E. degree in system design engineering and the M.E. and Ph.D. degrees in integrated design engineering from Keio University, Yokohama, Japan, in 2016, 2017, and 2020, respectively.
From 2019 to 2020, he was a Research Fellow with the Japan Society for the Promotion of Science.
Since 2020, he has been with Hiroshima University, Higashihiroshima, Japan.
His research interests include motion control, robotics, mechatronics, and control engineering.
\end{IEEEbiography}
\vspace{-10mm}
\begin{IEEEbiography}
[{\includegraphics[width=1in,height=1.25in,clip,keepaspectratio]{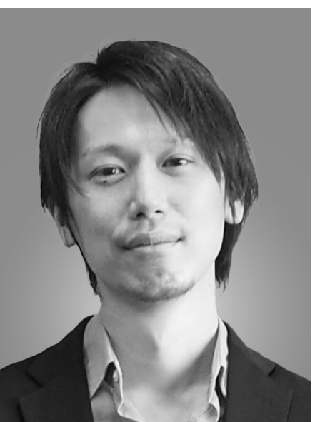}}]
{Yoshihiro Itaguchi}
received the Ph.D. degree in Psychology from Waseda University, Japan, in 2013. He is currently an Assistant Professor with the Psychology Department of Keio University. His current research interests include the interaction of body, movement, brain, and cognition in human.
\end{IEEEbiography}
\begin{IEEEbiography}
[{\includegraphics[width=1in,height=1.25in,clip,keepaspectratio]{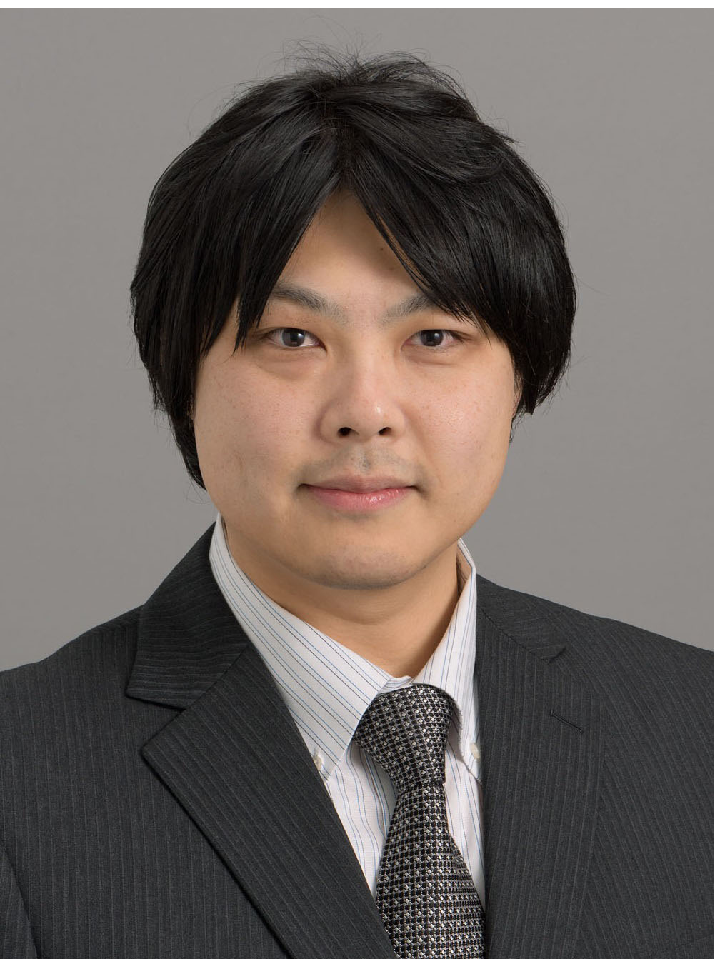}}]
{Seiichiro Katsura} (Member, IEEE)
received the B.E. degree in system design engineering and the M.E. and Ph.D. degrees in integrated design engineering from Keio University, Yokohama, Japan, in 2001, 2002 and 2004, respectively.

From 2003 to 2005, he was a Research Fellow of the Japan Society for the Promotion of Science (JSPS). From 2005 to 2008, he worked at Nagaoka University of Technology, Nagaoka, Niigata, Japan. Since 2008, he has been at Department of System Design Engineering, Keio University, Yokohama, Japan. Currently, he is working as a Professor. In 2017, he was a Visiting Researcher with the Laboratory for Machine Tools and Production Engineering (WZL) of RWTH Aachen University, Aachen, Germany. His research interests include applied abstraction, human support, data robotics, wave system, systems energy conversion, and electromechanical integration systems.

Prof. Katsura serves as Associate Editor of the IEEE Journal of Emerging and Selected Topics in Industrial Electronics and Technical Editor of IEEE/ASME Transactions on Mechatronics, and served as Associate Editor of the IEEE Transactions on Industrial Electronics. He was the recipient of The Institute of Electrical Engineers of Japan (IEEJ) Distinguished Paper Awards in 2003 and 2017, Yasujiro Niwa Outstanding Paper Award in 2004, The European Power Electronics and Drives-Power Electronics and Motion Control Conference, EPE-PEMC'08 Best Paper Award in 2008, IEEE Industrial Electronics Society Best Conference Paper Award in 2012, and JSPS Prize in 2016.
\end{IEEEbiography}

%% file: IEEE.bbl
\begin{thebibliography}{10}
\providecommand{\url}[1]{#1}
\csname url@samestyle\endcsname
\providecommand{\newblock}{\relax}
\providecommand{\bibinfo}[2]{#2}
\providecommand{\BIBentrySTDinterwordspacing}{\spaceskip=0pt\relax}
\providecommand{\BIBentryALTinterwordstretchfactor}{4}
\providecommand{\BIBentryALTinterwordspacing}{\spaceskip=\fontdimen2\font plus
\BIBentryALTinterwordstretchfactor\fontdimen3\font minus
  \fontdimen4\font\relax}
\providecommand{\BIBforeignlanguage}[2]{{%
\expandafter\ifx\csname l@#1\endcsname\relax
\typeout{** WARNING: IEEEtran.bst: No hyphenation pattern has been}%
\typeout{** loaded for the language `#1'. Using the pattern for}%
\typeout{** the default language instead.}%
\else
\language=\csname l@#1\endcsname
\fi
#2}}
\providecommand{\BIBdecl}{\relax}
\BIBdecl

\bibitem{Kumar2021}
S.~Kumar, C.~Savur, and F.~Sahin, ``Survey of human–robot collaboration in
  industrial settings: Awareness, intelligence, and compliance,'' \emph{{IEEE}
  Trans. Syst., Man, Cybern. Syst.}, vol.~51, no.~1, pp. 280--297, Dec. 2021.

\bibitem{Li2022}
Y.~Li, L.~Yang, D.~Huang, C.~Yang, and J.~Xia, ``A proactive controller for
  human-driven robots based on force/motion observer mechanisms,'' \emph{{IEEE}
  Trans. Syst., Man, Cybern. Syst.}, pp. 1--11, Jun. 2022.

\bibitem{Marvel2015}
J.~A. Marvel, J.~Falco, and I.~Marstio, ``Characterizing task-based
  human–robot collaboration safety in manufacturing,'' \emph{{IEEE} Trans.
  Syst., Man, Cybern. Syst.}, vol.~45, no.~2, pp. 260--275, Jul. 2015.

\bibitem{Liu2022}
M.~Liu, C.~Xiao, and C.~Chen, ``Perspective-corrected spatial referring
  expression generation for human-robot interaction,'' \emph{{IEEE} Trans.
  Syst., Man, Cybern. Syst.}, pp. 1--13, Apr. 2022.

\bibitem{Quintas2019}
J.~Quintas, G.~S. Martins, L.~Santos, P.~Menezes, and J.~Dias, ``Toward a
  context-aware human–robot interaction framework based on cognitive
  development,'' \emph{{IEEE} Trans. Syst., Man, Cybern. Syst.}, vol.~49,
  no.~1, pp. 227--237, May 2019.

\bibitem{Li2018}
H.-y. Li, I.~Paranawithana, L.~Yang, T.~Sey, K.~Lim, S.~Foong, F.~C. Ng, and
  U.-x. Tan, ``{Stable and Compliant Motion of Physical Human – Robot
  Interaction Coupled With a Moving Environment Using Variable Admittance and
  Adaptive Control},'' \emph{{IEEE} Trans. Robot. Autom. Letters}, vol.~3,
  no.~3, pp. 2493--2500, Jul. 2018.

\bibitem{Bowyer2015}
S.~A. Bowyer, F.~Rodriguez, and A.~Motivation, ``{Dissipative Control for
  Physical Human – Robot Interaction},'' \emph{{IEEE} Trans. Robot.},
  vol.~31, no.~6, pp. 1281--1293, Dec. 2015.

\bibitem{Hassan2018}
M.~Hassan, H.~Kadone, T.~Ueno, Y.~Hada, Y.~Sankai, and K.~Suzuki,
  ``{Feasibility of Synergy-Based Exoskeleton Robot Control in Hemiplegia},''
  \emph{{IEEE} Trans. Neural Syst. Rehabil. Eng.}, vol.~26, no.~6, pp.
  1233--1242, Jun. 2018.

\bibitem{Yu2015}
H.~Yu, S.~Huang, G.~Chen, Y.~Pan, and Z.~Guo, ``{Human – Robot Interaction
  Control of Rehabilitation Robots With Series Elastic Actuators},''
  \emph{{IEEE} Trans. Robot.}, vol.~31, no.~5, pp. 1089--1100, Oct. 2015.

\bibitem{Reed2008}
K.~B. Reed and M.~A. Peshkin, ``{Physical collaboration of human-human and
  human-robot teams},'' \emph{{IEEE} Trans. Haptics}, vol.~1, no.~2, pp.
  108--120, Jul.--Dec. 2008.

\bibitem{Amirshirzad2019}
N.~Amirshirzad, A.~Kumru, and E.~Oztop, ``{Human Adaptation to Human–Robot
  Shared Control},'' \emph{IEEE Trans. Human-Mach. Syst.}, vol.~PP, pp. 1--11,
  Apr. 2019.

\bibitem{Ficuciello2015}
F.~Ficuciello, L.~Villani, and B.~Siciliano, ``{Variable Impedance Control of
  Redundant Manipulators for Intuitive Human-Robot Physical Interaction},''
  \emph{{IEEE} Trans. Robot.}, vol.~31, no.~4, pp. 850--863, Aug. 2015.

\bibitem{Patton2016}
J.~L. Patton, ``{A Model for Human – Human Collaborative Object Manipulation
  and Its Application to Human – Robot Interaction},'' \emph{{IEEE} Trans.
  Robot.}, vol.~32, no.~4, pp. 880--896, Aug. 2016.

\bibitem{Peternel2017}
L.~Peternel, N.~Tsagarakis, and A.~Ajoudani, ``{A Human – Robot
  Co-Manipulation Approach Based on Human Sensorimotor Information},''
  \emph{{IEEE} Trans. Neural Syst. Rehabil. Eng.}, vol.~25, no.~7, pp.
  811--822, Jul. 2017.

\bibitem{Dimeas2016}
F.~Dimeas, S.~Member, and N.~Aspragathos, ``{Online Stability in Human-Robot
  Cooperation with Admittance Control},'' \emph{{IEEE} Trans. Haptics}, vol.~9,
  no.~2, pp. 267--278, Apr.--Jun. 2016.

\bibitem{Aydin2018}
Y.~Aydin, S.~Member, and O.~Tokatli, ``{Stable Physical Human-Robot Interaction
  Using Fractional Order Admittance Control},'' \emph{{IEEE} Trans. Haptics},
  vol.~11, no.~3, pp. 464--475, Jul.--Sep. 2018.

\bibitem{Shergill2003}
S.~S. Shergill, P.~M. Bays, C.~D. Frith, and D.~M. Wotpert, ``{Two eyes for an
  eye: The neuroscience of force escalation},'' \emph{Science}, vol. 301, no.
  5630, p. 187, Jul. 2003.

\bibitem{valles2013}
N.~L. Valles and K.~B. Reed, ``To know your own strength: overriding natural
  force attenuation,'' \emph{{IEEE} Trans. Haptics}, vol.~7, no.~2, pp.
  264--269, Apr.--Jun. 2013.

\bibitem{Bays2006}
P.~M. Bays, J.~R. Flanagan, and D.~M. Wolpert, ``{Attenuation of self-generated
  tactile sensations is predictive, not postdictive},'' \emph{PLoS Biology},
  vol.~4, no.~2, pp. 281--284, Jan. 2006.

\bibitem{takagi2016}
A.~Takagi, C.~Bagnato, and E.~Burdet, ``Facing the partner influences exchanges
  in force,'' \emph{Scientific reports}, vol.~6, p. 35397, Oct. 2016.

\bibitem{walsh2011}
L.~D. Walsh, J.~L. Taylor, and S.~C. Gandevia, ``Overestimation of force during
  matching of externally generated forces,'' \emph{The Journal of physiology},
  vol. 589, no.~3, pp. 547--557, Jan. 2011.

\bibitem{Onneweer2013}
B.~Onneweer, W.~Mugge, and A.~C. Schouten, ``{Human force reproduction error
  depends upon force level},'' in \emph{Proc. of the 2013 {IEEE} WHC}, Aug.
  2013, pp. 617--620.

\bibitem{Onneweer2016}
B.~Onneweer, W.~Mugge, and A.~C. Schouten, ``{Force Reproduction Error Depends
  on Force Level, whereas the Position Reproduction Error Does Not},''
  \emph{{IEEE} Trans. Haptics}, vol.~9, no.~1, pp. 54--61, Jan.--Mar. 2016.

\bibitem{1993_Murakami_FC}
T.~Murakami, F.~Yu, and K.~Ohnishi, ``Torque sensorless control in
  multidegree-of-freedom manipulator,'' \emph{{IEEE} Trans. Ind. Electron.},
  vol.~40, no.~2, pp. 259--265, Apr. 1993.

\bibitem{katsura2007modeling}
S.~Katsura, Y.~Matsumoto, and K.~Ohnishi, ``Modeling of force sensing and
  validation of disturbance observer for force control,'' \emph{{IEEE} Trans.
  Ind. Electron.}, vol.~54, no.~1, pp. 530--538, Feb. 2007.

\bibitem{2019_Sariyildiz_DOB}
E.~{Sariyildiz}, R.~{Oboe}, and K.~{Ohnishi}, ``Disturbance observer-based
  robust control and its applications: 35th anniversary overview,''
  \emph{{IEEE} Trans. Ind. Electron.}, vol.~67, no.~3, pp. 2042--2053, Mar.
  2020.

\bibitem{2015_Sariyildiz_DOB}
E.~Sariyildiz and K.~Ohnishi, ``Stability and robustness of
  disturbance-observer-based motion control systems,'' \emph{{IEEE} Trans. Ind.
  Electron.}, vol.~62, no.~1, pp. 414--422, Jan. 2015.

\bibitem{2015_Sariyildiz_FC}
E.~Sariyildiz and K.~Ohnishi, ``On the explicit robust force control via
  disturbance observer,'' \emph{{IEEE} Trans. Ind. Electron.}, vol.~62, no.~3,
  pp. 1581--1589, Mar. 2015.

\end{thebibliography}
